%% file: main.tex
\documentclass[12pt]{article}
\usepackage[utf8]{inputenc}
\usepackage[margin=1in]{geometry}
\usepackage{bm}
\usepackage{amsmath, color, natbib}
\usepackage{amssymb}
\usepackage{amsthm}
\usepackage{graphicx}
\usepackage{amsfonts}
\usepackage{mathrsfs}
\numberwithin{equation}{section}
\usepackage{booktabs}
\usepackage{caption}
\usepackage{subcaption}
\usepackage{float}
\usepackage{adjustbox, url}
\usepackage[table]{xcolor}
\usepackage[doublespacing]{setspace}
\usepackage{tikz}
\usepackage{authblk}
\usetikzlibrary{shapes,arrows,positioning}
\usepackage{algorithm}
\usepackage{algpseudocode}
\usepackage{changebar}
\usepackage{multirow}
\usepackage{hyperref}
\usepackage{subcaption} 

\usepackage[english]{babel}
\theoremstyle{definition}

\begin{document}

\title{\bf KM-GPT: An Automated Pipeline for Reconstructing Individual Patient Data from Kaplan–Meier Plots}
   \author{\small
    Yao Zhao$^{1,*}$, \quad Haoyue Sun$^{1,*}$, \quad Yantian Ding$^{1, \dagger}$, \quad Yanxun Xu$^{1}$\\
   \small $^1$ Department of Applied Mathematics and Statistics, Johns Hopkins University \\
   \small  $^*$ These authors contributed equally to this work\\
    \small  $\dagger$ Correspondence should be addressed to \texttt{yanxun.xu@jhu.edu}} 
    \date{}
  \maketitle

\begin{abstract}
Reconstructing individual patient data (IPD) from Kaplan–Meier (KM) plots provides valuable insights for evidence synthesis in clinical research. However, existing approaches often rely on manual digitization, which is error-prone and lacks scalability.  To address these limitations, we develop KM-GPT, the first fully automated, AI-powered pipeline for reconstructing IPD directly from KM plots with high accuracy, robustness, and reproducibility. KM-GPT integrates advanced image preprocessing, multi-modal reasoning powered by GPT-5, and iterative reconstruction algorithms to generate high-quality IPD without manual input or intervention.  Its hybrid reasoning architecture automates the conversion of unstructured information into structured data flows and validates data extraction from complex KM plots. To improve accessibility, KM-GPT is equipped with a user-friendly web interface and an integrated AI assistant, enabling researchers to reconstruct IPD without requiring programming expertise.  KM-GPT was rigorously evaluated on synthetic and real-world datasets, consistently demonstrating superior accuracy. To illustrate its utility, we applied KM-GPT to a meta-analysis of gastric cancer immunotherapy trials, reconstructing IPD to facilitate evidence synthesis and biomarker-based subgroup analyses. By automating traditionally manual processes and providing a scalable, web-based solution, KM-GPT transforms clinical research by leveraging reconstructed IPD to enable more informed downstream analyses, supporting evidence-based decision-making.
\end{abstract}


\section{Introduction}
Time-to-event data is a fundamental data type across medicine, public health, and social sciences. From studying patient survival and disease recurrence to evaluating the duration of unemployment, these analyses rely on detailed event timing information to draw meaningful conclusions. While individual-level event data provides the most comprehensive information for analysis, which enables precise modeling, subgroup identification, and assumption validation, access to such data is often restricted. In practice, researchers frequently face a major barrier: pharmaceutical companies and research consortia routinely publish Kaplan-Meier (KM) survival plots in clinical reports but rarely share the underlying individual patient data (IPD) due to privacy concerns, proprietary interests, or regulatory constraints. As a result, systematic reviews often depend solely on published data, which is typically restricted to KM  curves and a limited set of summary statistics. This dependence on aggregated data poses significant challenges for secondary analyses. For instance, without IPD, researchers cannot validate proportional hazards assumptions or examine treatment effects in patient subgroups. Recognizing these challenges, a growing body of research has focused on reconstructing IPD from KM plots for secondary analyses and evidence synthesis \cite{dear1994iterative, arends2008meta, fiocco2009meta, parmar1998extracting, ouwens2010network, jansen2011network, earle2000assessment, williamson2002aggregate}.

Reconstructing IPD from KM plots generally involves two key steps: digitizing the graph and reconstructing survival data. Digitizing graphs involves extracting coordinate values, typically time and survival probability, from static image plots of KM curves. This step transforms visual information into quantitative data but often requires substantial preprocessing and manual interaction. Several software tools such as DigitizeIt \cite{DigitizeIt}, ScanIt \cite{ScanIt}, and PlotDigitizer \cite{PlotDigitizer} allow users to import graph images, calibrate axes, and manually or semi-automatically extract data points from KM curves. While effective, these tools require significant manual input, such as setting axis ranges and scaling, and performing point-and-click operations to select data points. This process can be time-consuming and prone to human error. More recently, SurvdigitizeR \cite{zhang2024survdigitizer} introduced a scripting-based approach to partially automate the digitization process. However, it requires users to manually input x- and y-axis ranges, specify tick marks, and often preprocess figures to ensure they are correctly formatted. These manual steps demand expertise and introduce potential inconsistencies, further underscoring the need for fully automated solutions that reduce human effort and improve reproducibility.

Once the digitization step is complete, the next step is data reconstruction, which involves generating IPD from the digitized KM curves. Guyot et al. \cite{guyot2012enhanced} developed an iterative algorithm, often referred as the iKM algorithm, to reconstruct IPD from KM curves. This approach uses digitized KM data combined with supplementary information such as the number of patients at risk and total events to approximate the original IPD. Building on the iKM method, Liu et al. \cite{liu2021ipdfromkm} introduced IPDfromKM, a two-stage workflow that integrates curve digitization and IPD reconstruction into a single pipeline. While IPDfromKM demonstrates significant improvements in efficiency and ease of use, it still requires substantial manual input, limiting the automation and scalability of the process.  Specifically, IPDfromKM requires users to manually click on the KM curve to extract survival coordinates, making the task labor-intensive and susceptible to inaccuracies. Additionally, the number-at-risk table must be manually entered, as the tool cannot automatically extract this information from  graphical figures. Furthermore, users are required to manually specify the tick values or scales for the x-axis (typically time), which involves interpreting the figure and risks introducing inaccuracies if calibration is imprecise.

To address these limitations, we develop KM-GPT, a fully automated end-to-end pipeline for reconstructing IPD from KM  plots. KM-GPT combines advanced image processing techniques with multi-modal AI reasoning powered by the state-of-the-art large language model GPT-5~\cite{openai2025gpt5} to overcome the challenges posed by visual and semantic variability in KM plots, such as differences in styles, layouts, and resolutions. The pipeline processes KM plots by validating their structure, extracting graphical and textual components, including axis labels, risk tables, and survival curves through GPT-5’s contextual reasoning to resolve ambiguities like censoring indicators and overlapping curves. These intermediate results are then seamlessly integrated to produce structured IPD that preserves the statistical properties of the original KM plots. By minimizing user intervention, handling complex layouts and figure heterogeneity, and ensuring reproducibility, KM-GPT provides a scalable solution for generating IPD datasets suitable for downstream analyses such as survival data modeling and systematic meta-analysis.

The novelty of KM-GPT lies in its end-to-end automation, robust preprocessing, and multi-modal reasoning capabilities, which collectively address the longstanding challenges of extracting and reconstructing IPD from KM plots. Our contributions are fourfold. First, KM-GPT introduces the first fully automated pipeline for IPD reconstruction, eliminating the need for labor-intensive manual steps such as digitizing curve points, calibrating axes, or transcribing textual data from figures. 
Second, KM-GPT pioneers the application of multi-modal reasoning in the context of KM plots by combining optical character recognition (OCR) techniques with GPT-5’s advanced contextual and cross-modal reasoning capabilities. It seamlessly interprets both textual and graphical inputs, resolving ambiguities from low-quality visual features, reducing uncertainty by cross-validating outputs based multi-modal sources, and ensuring that key clinical data are extracted accurately and preserved in the final output. 
Third, KM-GPT introduces an automated mechanism for transforming noisy piecewise information including axis tick labels and embedded risk tables into well-formatted, machine-readable intermediate data. By leveraging large language model's structural output capabilities~\cite{wang2025slot}, the system aggregates and organizes piecewise data to JSON format with strictly consistent predefined structure.
Finally, we design a user-friendly web interface that enables researchers to reconstruct IPD without requiring statistical expertise or specialized programming skills.


The remainder of this paper is organized as follows. Section~\ref{sec:methods} describes the proposed KM-GPT pipeline and its core components. Section~\ref{sec:interface} introduces the user-friendly web interface developed for KM-GPT. In Section~\ref{sec:simulation}, we evaluate KM-GPT's performance through extensive synthetic data studies, demonstrating its robustness and accuracy under diverse trial-like conditions. Section~\ref{sec:realdata} assesses its performance on manually labeled real-world datasets. We demonstrate the downstream utility of KM-GPT in a meta-analysis of gastric cancer immunotherapy trials in Section~\ref{sec:meta}. Finally, Section~\ref{sec:Discussion} concludes the paper with a discussion.

 

\section{Methods}
\label{sec:methods}

\begin{figure}[htbp!]
    \centering
    \includegraphics[width=0.95\linewidth]{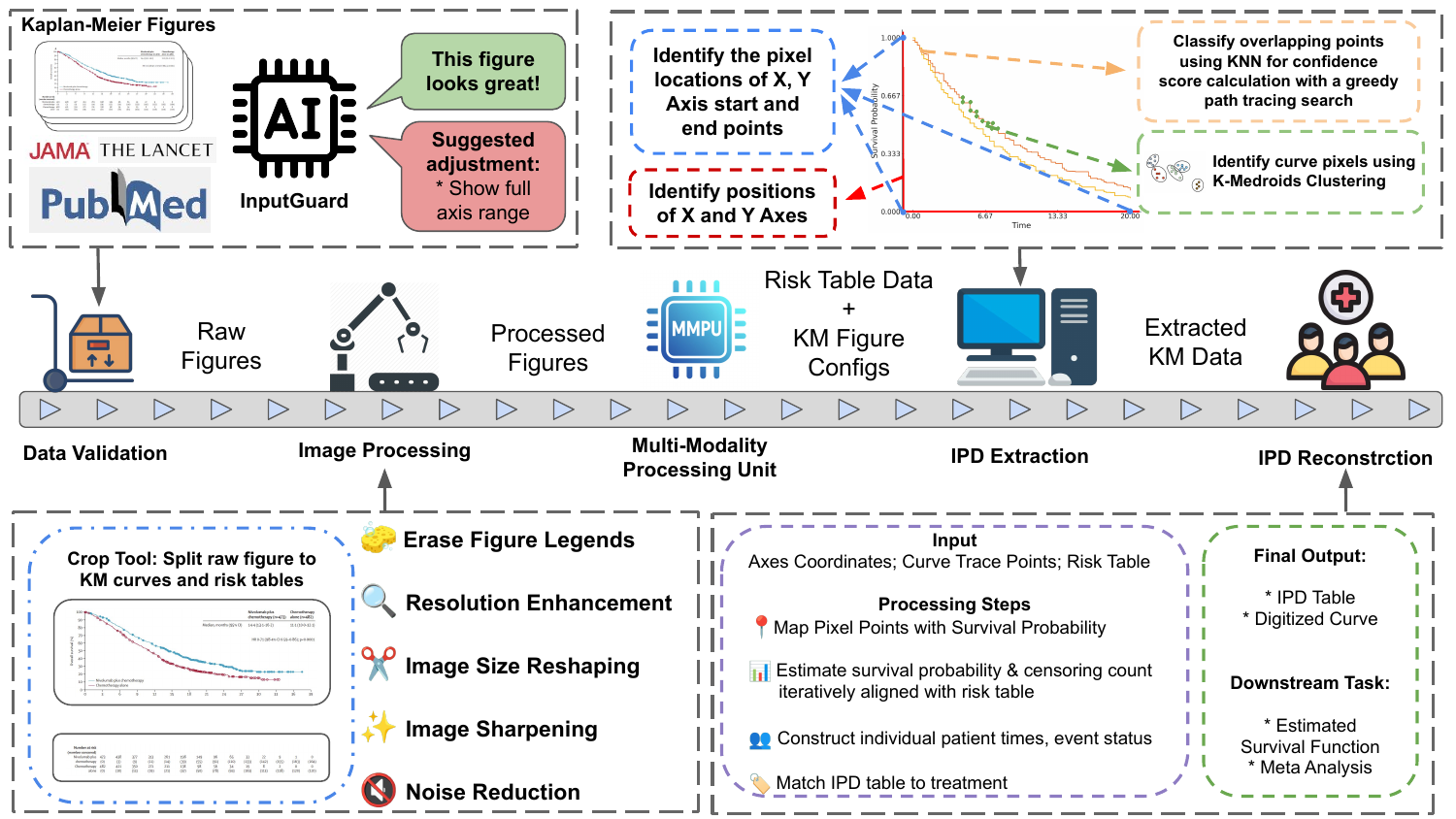}
    \caption{Overview of the KM-GPT pipeline.}
    \label{fig:Method Working Structure}
\end{figure}

Figure~\ref{fig:Method Working Structure} presents the end-to-end pipeline of KM-GPT for reconstructing IPD from KM plots. The framework integrates image processing techniques and multi-modal reasoning with GPT-5 to achieve full automation, adaptability, and reproducibility. KM-GPT is organized into five functional modules: \textbf{Data Validation}, \textbf{Image Processing}, \textbf{Multi-Modality Processing Unit (MMPU)}, \textbf{IPD Extraction}, and \textbf{IPD Reconstruction}. Each module addresses a distinct stage in the transformation from raw KM plots to structured datasets. The design and implementation of these modules are detailed below.

\subsection{Data Validation and Image Processing}

The Data Validation and Image Processing modules form the preprocessing backbone of KM-GPT, enabling reliable standardization and analysis of KM plots that originate from a wide array of sources and often vary in layout, resolution, and stylistic conventions.

The pipeline begins with Data Validation, which inspects input KM plots for completeness and suitability for automated processing. Ensuring high-quality inputs is essential to minimize downstream errors and improve reproducibility, especially given the heterogeneity in figure formats and presentation styles. The core of this module is InputGuard, an AI diagnostic agent powered by GPT-5. InputGuard evaluates both the structural and semantic integrity of KM plots through tasks such as object detection, layout parsing, and image–text alignment. Using structured prompts, InputGuard simultaneously checks key quality requirements for successful KM-GPT processing. Specifically, InputGuard verifies the presence of essential components, including axis labels, tick marks, survival curves, legends, and risk tables. When deficiencies or ambiguities are identified (e.g., missing axis ticks or incomplete risk tables), the system generates natural language feedback with specific recommendations for correction before the figure advances to downstream modules. This built-in feedback loop ensures that inconsistencies are resolved in early stages to prevent error propagation. In addition to validation, InputGuard serves as a standardization gateway. By normalizing heterogeneous inputs and enforcing a minimum visual–semantic quality threshold, it ensures smooth integration of figures into the KM-GPT pipeline. This process minimizes user-side debugging and significantly increases the success rate of automated IPD reconstruction across diverse figure formats.

Once validated, figures proceed to the Image Processing module, which transforms raw KM plots into images with standardized quality optimized for automated feature extraction. While this module operates fully automatically, users have the option to interactively refine inputs for greater precision. For instance, users can crop figures into individual KM curve panels and risk tables or manually remove distracting elements, such as legends, watermarks, or annotations, that might obstruct curve tracing or text recognition based on empirical evaluations.

Once user confirms customized input images, KM-GPT applies a sequence of automated enhancement steps. First, image resolution is improved using the efficient sub-pixel convolutional neural network (ESPCN) model~\cite{shi2016real}, implemented via OpenCV~\cite{opencv_library}.
This operation enhances the image resolution by a factor of 2, enabling finer detection of features such as axis ticks, text labels, and curve points that might otherwise be lost in low-resolution plots. The plot is then resized to a predefined dimension for computational efficiency while preserving detail. Next, edge clarity is enhanced with a Laplacian kernel–based sharpening filter~\cite{gonzalez2008digital}, which improves delineation of structural elements such as survival curves and axis boundaries. In the context of KM plots extraction, sharpening is crucial since visual elements like step-wise survival curves and small textual labels are often thin and faint. It amplifies the visibility of these high-frequency features without introducing significant distortion, thereby improving the performance of both visual parsing and digitization steps in the downstream pipeline. Finally, denoising is performed using the non-local means denoising algorithm~\cite{buades2005non} to reduce background noise while retaining critical information. This step is particularly useful in scanned documents or plots extracted from low-quality PDFs, where compression artifacts and visual clutter may obscure essential content. 

Together, these modules combine automated cleaning and enhancement with human-in-the-loop customization, offering both flexibility and scalability. By ensuring input quality and visual consistency, they establish a solid foundation for accurate curve interpretation and risk table parsing in downstream stages.

\subsection{Multi-Modality Processing Unit (MMPU)}
\begin{figure}[htbp!]
    \centering
    \includegraphics[width=0.95\linewidth]{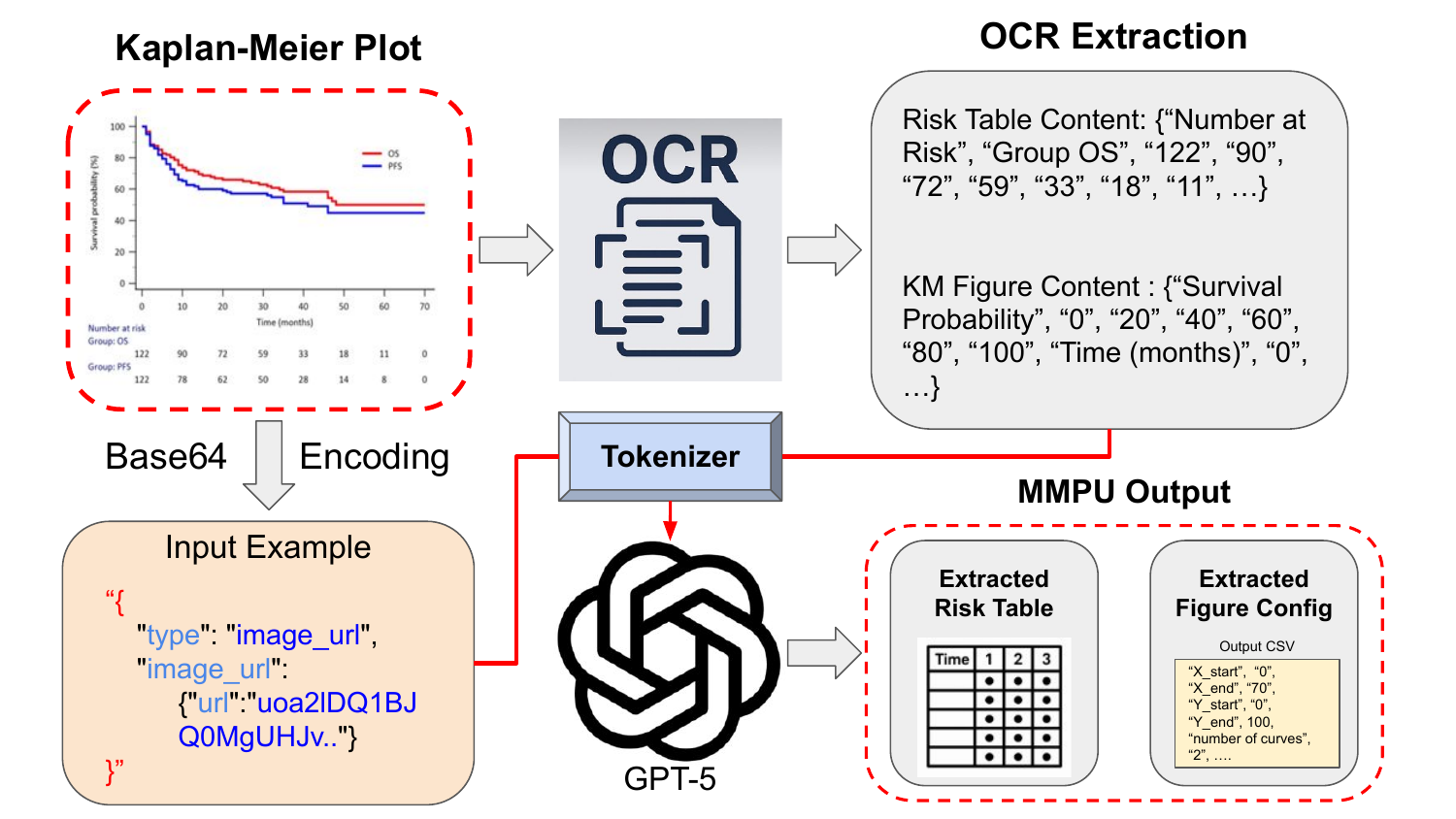}
    \caption{Demonstration of Multi-Modal Processing Unit (MMPU)}
    \label{fig:MMPC Structure}
\end{figure}

The MMPU module is a core innovation of KM-GPT, designed to convert non-structured inputs including KM plots and their associated risk tables into structured data, a process that traditionally requires extensive manual effort. As illustrated in Figure~\ref{fig:MMPC Structure}, MMPU introduces a hybrid architecture that fuses classical optical character recognition (OCR) with modern multi-modal AI reasoning powered by GPT-5. This design enables the system to reliably handle the visual heterogeneity and semantic ambiguity that characterize KM plots, especially in publications with complex layouts or non-standard conventions.

The MMPU operates in two integrated stages. In the first stage, a high-resolution OCR engine parses textual elements embedded in the KM plot, including axis tick labels, time intervals, curve annotations, and risk table entries. These extracted tokens, converted from raw pixel-level patterns into textual units, are organized into semantically grouped regions, bridging the gap between non-structured inputs and structured intermediate representations. To improve robustness across diverse figure styles and layouts, we employ two complementary image processing strategies, each tailored to the unique visual characteristics of risk table regions and axis label regions. For risk tables, we adopt adaptive thresholding with Gaussian weighting~\cite{opencv_library}~\cite{opencv_threshold} to extract structured numeric data. This technique computes a local threshold for each pixel based on a weighted sum of neighboring intensities, with greater emphasis on the center of the window. By adapting locally, the method effectively binarizes both prominent and faint text, substantially improving OCR performance. For axis label regions, we instead apply global thresholding~\cite{opencv_library}~\cite{opencv_cv_threshold_ref} with a fixed threshold value. Axis labels are typically printed in high contrast against uniform backgrounds and are spatially isolated from other graphical elements, making global thresholding sufficiently accurate and computationally efficient. KM-GPT also employs different OCR engine settings to accommodate different tasks, with details reported in Supplementary Material Section~\ref{sec:OCR_Engine_setting}. This approach avoids unnecessary processing while preserving the clarity of axis tick labels, ensuring reliable OCR without added complexity. Together, these strategies ensure that heterogeneous visual text regions are converted into clean, machine-readable tokens.

In the second stage, GPT-5 processes these OCR outputs with the original image through its unified multi-modal attention architecture. The visual input is converted into base-64 encoding, which represents image data as ASCII characters, and passed to the model alongside OCR-derived text embeddings. Within this unified representation, the model performs semantic and visual reasoning jointly to resolve KM-specific ambiguities, such as identifying group labels or confidence intervals. For example, it detects group labels by aligning curve shapes and colors with numbers at risk and cross-validates these findings against extracted textual annotations. By combining rule-based precision from OCR with contextual reasoning from GPT-5, MMPU achieves high-fidelity interpretation of KM-specific features that are often lost when handling non-structured data using vision or language models alone. The final output of MMPU is a regularized JSON file containing necessary parameters, risk tables, and relevant metadata. Crucially, this structured representation preserves the data structure for downstream modules, ensuring that the extracted data is not only accurate but also directly suitable for IPD reconstruction steps.

In summary, MMPU transforms non-structured KM plots into structured intermediate model parameters through a reproducible, automated pipeline. By bridging classical vision methods with state-of-the-art multi-modal reasoning, it provides a powerful and generalizable solution for extracting high-fidelity meta information from heterogeneous KM plots. 

\subsection{IPD Extraction and Reconstruction}
Following the extraction of curve parameters, axis configurations, and number-at-risk tables by the MMPU, the IPD Extraction and Reconstruction modules perform the task of converting visual survival curves into calibrated IPD. By systematically integrating pixel-level curve coordinates with survival probability mappings, these modules produce structured, analysis-ready data aligned to the temporal and probabilistic scales derived in earlier stages.

The process begins with axis localization and pixel-to-scale calibration. During preprocessing, the detected start and end coordinates of both axes define a linear transformation from image pixel space to the time (x-axis) and survival probability (y-axis) domains, ensuring precise quantification of each traced curve point in real-world units. Curve isolation and path tracing are then performed using a K-medoids clustering approach to group curve pixels, followed by greedy search path tracing to reconstruct the ordered trajectory of each survival curve. For publications with overlapping or visually entangled curves, the k-nearest neighbors ($k$-NN) classification is applied to assign pixels to the correct group, with confidence scores indicating assignment certainty. These scores are subsequently used to refine point ordering and suppress spurious connections. Once survival coordinates are calibrated, the iterative event reconstruction algorithm from iKM~\cite{guyot2012enhanced} is adapted to align survival probabilities with the extracted number-at-risk table. This step iteratively reconciles event and censoring assignments to ensure reconstructed risk sets match reported values, while preserving the shape of the digitized survival curve. Additional technical details on the calibration and reconstruction modules can be found in Supplementary Material Section \ref{sec:Method Details}.

Finally, a reconstructed IPD table is generated, containing time, event indicators (1 = event, 0 = censored), and treatment or group labels inferred from the curve source. To ensure high reconstruction fidelity, the reconstructed IPD is immediately validated by re-plotting KM curves from the model output and comparing them to the originally extracted curves. The resulting CSV-formatted IPD is readily usable for downstream applications, including survival analysis, subgroup identification, and evidence synthesis.

\section{KM-GPT Interface}
\label{sec:interface}
\begin{figure}[htbp!]
    \centering
    \begin{tikzpicture}
        \node[anchor=center] (main) at (0, 10) {\includegraphics[width=0.95\linewidth]{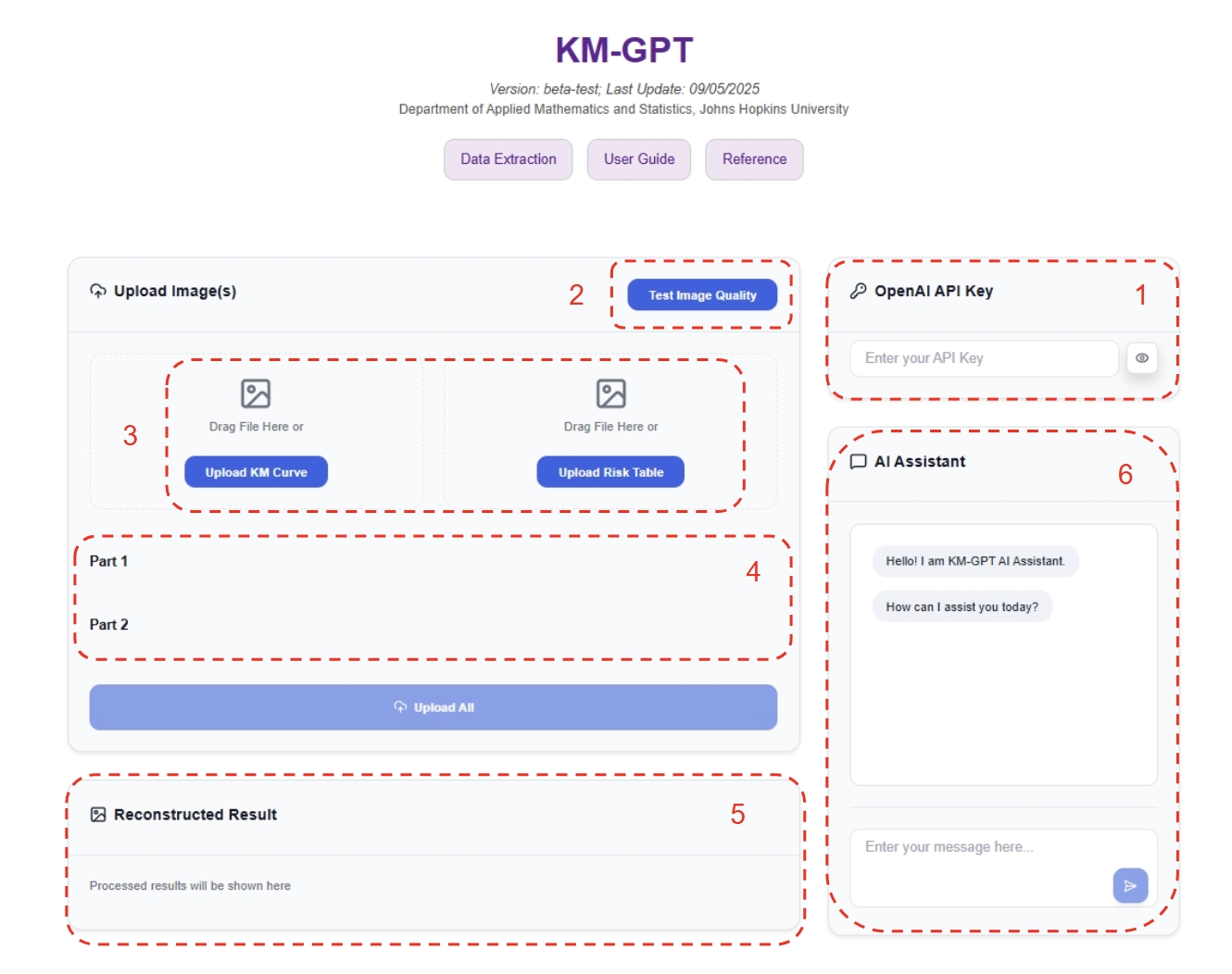}}; 

        \node[anchor=center] (left) at (-4, 0) {\includegraphics[width=0.4\linewidth]{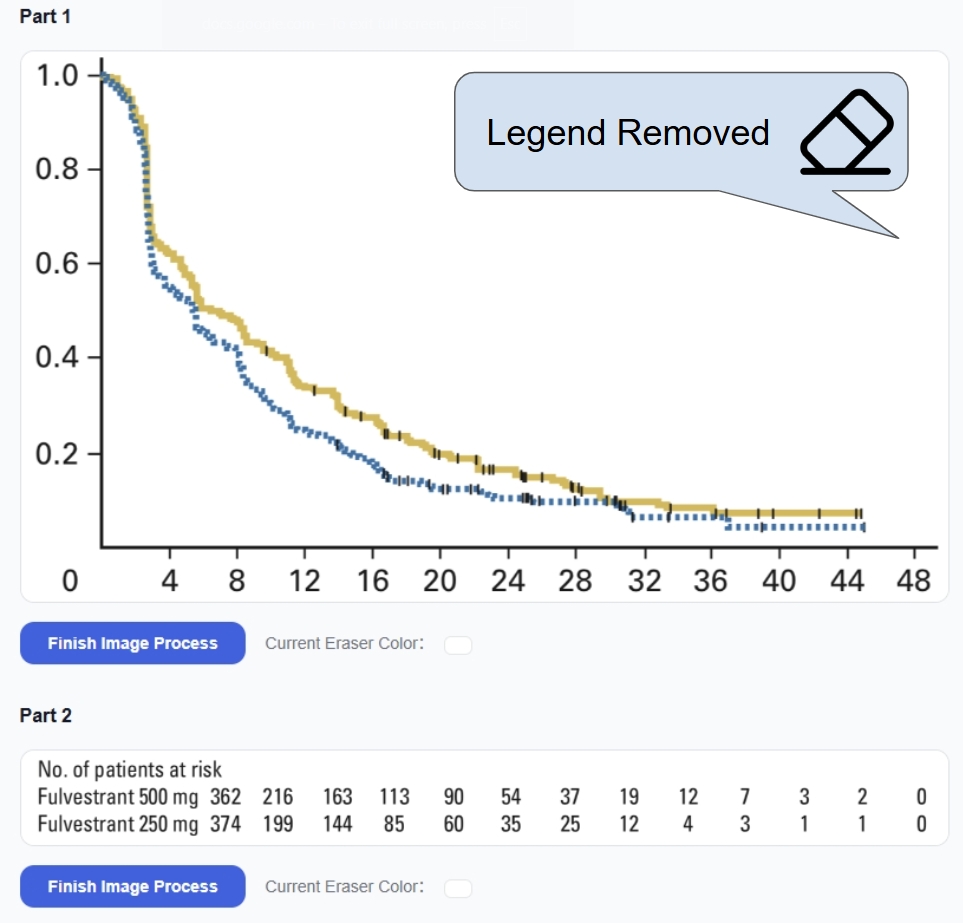}};
    
        \node[anchor=center] (right) at (4, 0) {\includegraphics[width=0.54\linewidth]{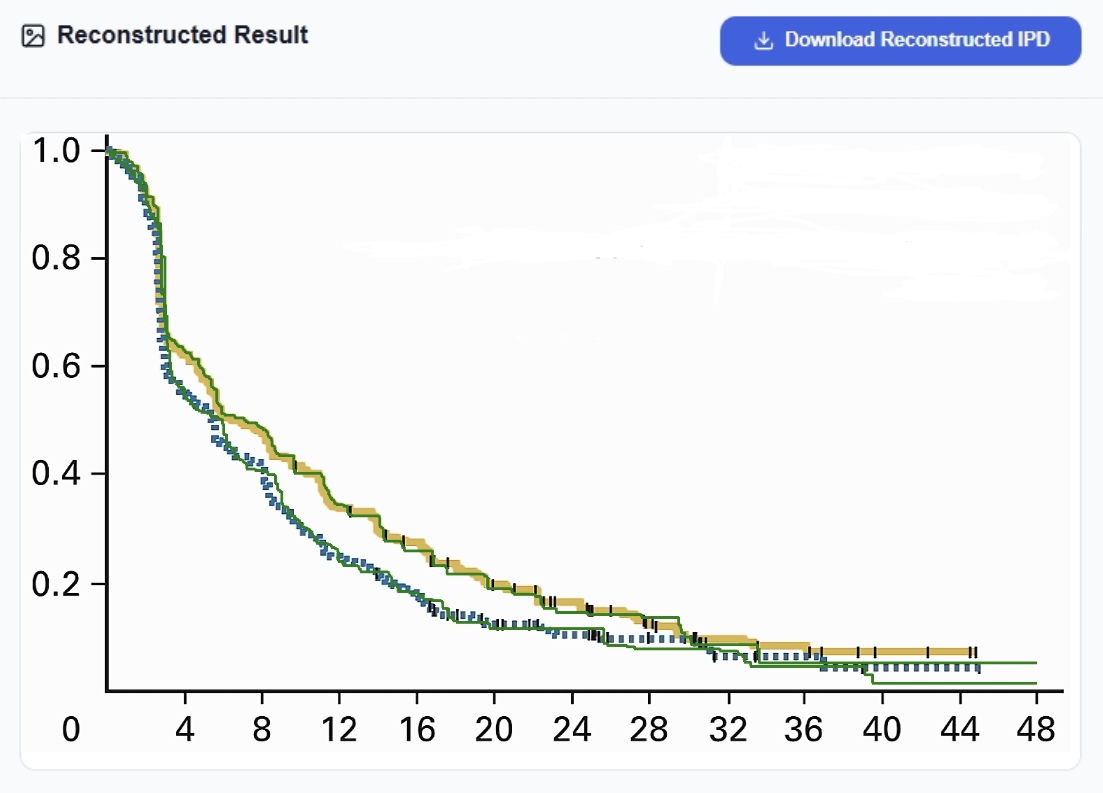}};

        \draw[->, ultra thick, blue] (0, 8) -- (-2.5, 3);
        \draw[->, ultra thick, red] (0, 5) -- (2.5, 3);
    \end{tikzpicture}
    \caption{KM-GPT Website Tool Page.}
    \label{fig:tool_main_page}
\end{figure}

To make KM-GPT accessible to a broader audience beyond command-line or script-based workflows, we develop a user-friendly web-based interface (\url{https://km-gpt.wse.jhu.edu/}) designed specifically for researchers and clinicians with limited programming experience. The web interface follows a streamlined upload–process–output workflow, emphasizing simplicity and usability with minimal configuration. Figure~\ref{fig:tool_main_page} illustrates the KM-GPT tool, with key modules highlighted in red boxes.

Box 1 allows users to input their OpenAI API keys, which are securely stored in local browser cookies and never saved by the interface. Box 2 provides preprocessing feedback powered by InputGuard, verifying axis labels, risk tables, and overall figure quality to ensure the uploaded images meet the processing requirements. Image uploads are handled in Box 3, where users can upload KM plots either by selecting files from their local computer or through drag-and-drop functionality. After uploading, users are directed to the image preparation stage in Box 4 (with a zoomed-in view shown in the lower-left portion of Figure~\ref{fig:tool_main_page}, indicated by the blue arrow). In this stage, tools for cropping and cleaning figures are provided, including an internal eraser for noisy elements removal. Once the preparation is complete, reconstructed survival curves are presented in Box 5, overlaid with the original KM plots for validation. This section also includes a downloadable reconstructed IPD dataset, enabling further analysis of the processed data.

Additionally, KM-GPT integrates a dedicated agent (Box 6) within the interface to assist users with task-specific troubleshooting and guidance. The agent is powered by a retrieval-augmented generation (RAG) system based on LMAR~\cite{zhao2025lmar}, a RAG augmentation method designed for private knowledge adaptation, which incorporates user guides, developer documentation, and troubleshooting notes to enhance awareness of KM-GPT’s backend workflow. This agent is connected to input images, user queries, processing logs, and reconstructed outputs, providing contextual support for each task. To maintain focus and efficiency, the agent’s assistance is restricted to model-related queries and limited conversation rounds, ensuring concise and task-specific guidance.




    


\section{Performance Evaluation on Synthetic Data}
\label{sec:simulation}
To evaluate the performance of KM-GPT in reconstructing IPD from KM plots, we first conducted simulation studies using synthetic data to systematically test its accuracy and robustness under varying sample sizes, survival durations, and censoring rates. These scenarios were designed to reflect the variability typically encountered in survival analysis. For each synthetic dataset, KM curves and corresponding risk tables were generated to summarize survival probabilities and the number of individuals at risk at predefined time points. KM-GPT was then applied to these plots to reconstruct the underlying IPD and assess its performance across the simulated scenarios. 

\subsection{Generation of Synthetic Data}
For each synthetic dataset, we assumed a sample size of $n$. The survival time $T_i$ for each individual $i$ was independently sampled from an exponential distribution, $T_i\sim \mathrm{Exp}(\lambda)$, where the median survival time can be computed as $\log(2)/\lambda$. To incorporate right censoring, we specified a target censoring rate $\eta$, and a fixed proportion of patients ($n\times \eta$) was randomly selected for censoring. For each censored patient, the censoring time $C_i$  was sampled uniformly from the interval $[0, \min(T_i, \tau)]$, where $\tau$ is the maximum follow-up time. The observed survival time for each individual was then defined as $Y_i=\min(T_i, C_i)$, with the censoring indicator $\delta_i=I(T_i\leq C_i)$, where $I(\cdot)$ is the indicator function.

To reflect realistic clinical scenarios, we varied three key input parameters: sample size, median survival time, and censoring rate. For each simulation, these parameters were sampled from normal distributions with distinct means and standard deviations corresponding to three levels: low, medium, and high, as summarized in Table~\ref{tab:simulation-settings}.  For example, in one scenario representing a medium-sized study with a long median survival time and a medium censoring rate, the sample size was drawn from a normal distribution with a mean of 200 and a standard deviation of 30. Similarly, the median survival time was sampled from a normal distribution with a mean of 36 months and a standard deviation of 6 months, while the censoring rate was drawn from a normal distribution with a mean of 0.3 and a standard deviation of 0.05.

\begin{table}[htbp!]
\centering
\caption{Parameter Settings for Generating  Synthetic Datasets. For each parameter (sample size, median survival time, and censoring rate), values were sampled from normal distributions with specified means and standard deviations ($\mu$, $\sigma$). }
\begin{tabular}{lccc}
\toprule
\textbf{Parameter} & \textbf{Low} & \textbf{Medium} & \textbf{High} \\
\midrule
Number of Patients & (50,\ 10) & (200,\ 30) & (800,\ 50) \\
Median Survival Time (months) & (6,\ 1) & (12,\ 2) & (36,\ 6) \\
Censoring Rate & (0.05,\ 0.02) & (0.3,\ 0.05) & (0.7,\ 0.08) \\
\bottomrule
\end{tabular}
\label{tab:simulation-settings}
\end{table}

These combinations of parameter values were used to simulate diverse trial conditions. Small cohorts with short survival times and high censoring rates were designed to replicate certain early-phase studies or trials for rare diseases, while large cohorts with long survival times and low censoring rates were used to emulate long-term studies conducted in preventive medicine or chronic conditions. To ensure robustness, we generated 20 replicates for each parameter combination, resulting in $3\times 3\times 3=27$ unique parameter combinations and a total of $20\times 27=540$ KM curves with their associated risk tables.

Supplementary Figure~\ref{fig:simulated_KM_figures} provides examples of simulated KM plots and their corresponding risk tables. Specifically, the left panel illustrates a KM plot for a scenario with a small number of patients, producing a step-like survival curve with large vertical drops and unstable tail estimates due to the high variability typical in small and heavily censored cohorts. In contrast, the right panel shows a KM plot for a much larger trial, where the survival curve declines smoothly and continuously, exhibiting stable tail behavior and narrower steps, reflecting the reliability of estimates in large cohorts with low censoring rates.

\subsection{Results on Synthetic Data}

The evaluation of KM-GPT was performed on all 540 KM plots generated from the synthetic datasets. First, we evaluated the success rate of the image processing pipeline and assessed the accuracy of automatically extracted parameters, including axis ranges, increments, and risk table values. Second, we compared the survival curves reconstructed by KM-GPT to the simulation ground truth to evaluate the fidelity and accuracy of the final IPD reconstruction.

\subsubsection{Parameter Extraction Success and Accuracy}
Data extraction after image processing represent the foundational steps of the KM-GPT pipeline, ensuring that digitized KM curves and risk tables can be extracted effectively. KM-GPT successfully processed 538 of the 540 KM plots, yielding an overall image processing success rate of 99.6\%. The two plots where processing failed were attributed to digitization errors, where survival curves or axis information were either processed incorrectly or extracted incompletely. These errors caused mismatches between hyperparameters extracted from the MMPU and values detected through OCR, leading to digitization failures.

For the 538 successfully processed plots, we next evaluated the accuracy of automatically extracted KM plot parameters, including axis ranges, tick intervals, and the risk table values. These extracted values were compared against the simulation ground truth. KM-GPT achieved a remarkable accuracy of 100\% for axis ranges (x\_start, y\_start, x\_end, y\_end) and tick intervals, ensuring that all visual components defining the KM curve scales (time and survival probability) aligned perfectly with the known ground truth. Risk table extraction, which involves the complex task of transcribing the number-at-risk values at predefined time points, was also performed with 100\% accuracy. These results demonstrate the system’s robustness across all critical digitization tasks essential for reliable reconstruction of IPD.


To compare KM-GPT's performance with existing tools, we evaluated its results against SurvdigitizeR, an R package designed to digitize KM curves. Unlike KM-GPT, SurvdigitizeR does not process risk tables, an important feature of KM plots that is essential for accurate reconstruction of IPD. As SurvdigitizeR lacks support for risk table extraction, users are required to manually remove these tables from figures before analysis, adding extra preprocessing steps and reducing the tool's applicability to raw KM plots. For this evaluation, we prepared ``cleaned" KM plots by manually removing the risk tables to ensure compatibility with SurvdigitizeR. Out of the 540 KM plots, SurvdigitizeR failed to process 178 figures, resulting in a failure rate of 33.0\%. Specifically, 165 plots encountered digitization errors, while 13 plots failed due to image formatting and processing issues. These failures stemmed from difficulties in selecting consistent manually-tuned image processing hyperparameters (e.g., background brightness, word sensitivity, and OCR version) in SurvdigitizeR for all figures. By contrast, KM-GPT eliminates this issue through its unified image preprocessing pipeline in the Image Processing module, enabling robust handling of diverse KM plot formats.

\subsubsection{Reconstruction of Survival Curves Evaluation}
We evaluated the accuracy of reconstructed IPD by comparing survival curves estimated from the KM-GPT output to the ground truth survival curves derived from the simulated data. This analysis focused on the 538 KM plots successfully processed by KM-GPT.


We first measured point-wise reconstruction accuracy by calculating the absolute error (AE) between the survival probabilities estimated from the reconstructed IPD and those obtained from the simulated ground truth. Time was normalized to the interval [0,1] to ensure comparability across datasets with varying follow-up durations. The results, visualized in Supplementary Figure~\ref{fig:SimulationAE}, demonstrate that KM-GPT achieved consistently low reconstruction errors throughout the follow-up period. Across all simulations, the median AE was 0.005 (95\% CI: 0.000–0.034), underscoring KM-GPT's precision in estimating survival probabilities at every time point. Notably, we observed the variance of reconstruction error increases during the later follow-up periods, particularly after 75\% of the maximum follow-up time. Two main factors contribute to this pattern. First, the tail regions of survival curves naturally contain fewer observed events, since the risk set diminishes sharply as patients are censored or experience the event earlier. 
Second, in the tail regions, survival curves sometimes run very close to the time axis, where curves pixels are misidentified as axis pixels. Under these conditions, the digitization algorithm struggles to capture curve points with high precision, leading to greater variability in reconstructed estimates.

Next, we calculated the integrated absolute error (IAE) to evaluate deviation across the entire follow-up period. The IAE quantifies the area between the true survival curve and the reconstructed curve over normalized time and is defined as:
\begin{equation*}
    \mathrm{IAE} \;=\; \int_0^1 \big| S_{\text{truth}}(t) - S_{\text{reconstructed}}(t) \big| \, dt,
\end{equation*}
where $S_{\text{truth}}(t)$ and $S_{\text{reconstructed}}(t)$ denote the survival probabilities estimated from the ground truth and reconstructed IPD, respectively. Smaller IAE values indicate closer alignment between the curves, with 0 reflecting perfect agreement. Across the 538 datasets, KM-GPT achieved a median IAE of 0.018 (95\% CI: 0.002–0.088), demonstrating its overall accuracy in reconstructing survival curves.

To further assess the fidelity of reconstructed IPD, we also estimated the median overall survival (OS), a clinically important measure representing the time at which 50\% of the population is expected to survive. We computed the AE in median OS as the difference between reconstructed and true median survival times. Under normalized time settings, an AE in median OS of 0.001 corresponds to a deviation of 1 day per 1000 days of maximum follow-up. KM-GPT achieved a near-zero AE in median OS of 0.005 (95\% CI: 0.000–0.088).

\subsubsection{Performance Analysis Across Parameter Settings}
\begin{figure}[ht!]
    \centering
    \renewcommand{\thesubfigure}{\Alph{subfigure}}
    \begin{subfigure}{0.95\linewidth}
        \centering
        \includegraphics[width=\linewidth]{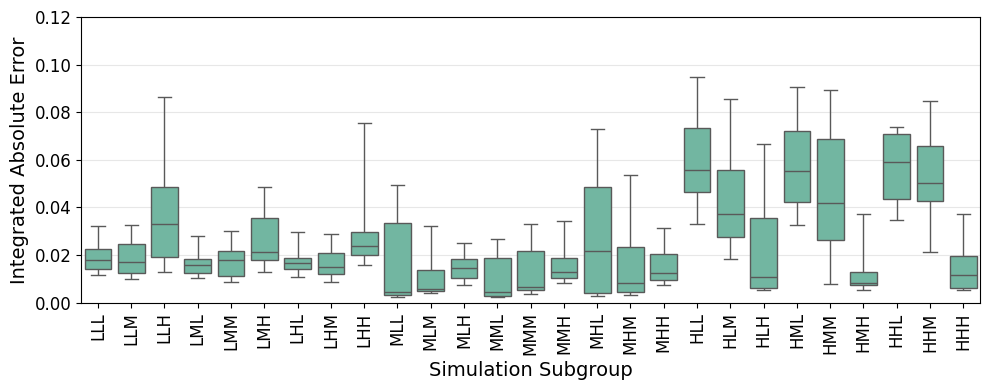}
        \caption{}
        \label{fig:SimulationMAEBox}
    \end{subfigure}
    \begin{subfigure}{0.95\linewidth}
        \centering
        \includegraphics[width=\linewidth]{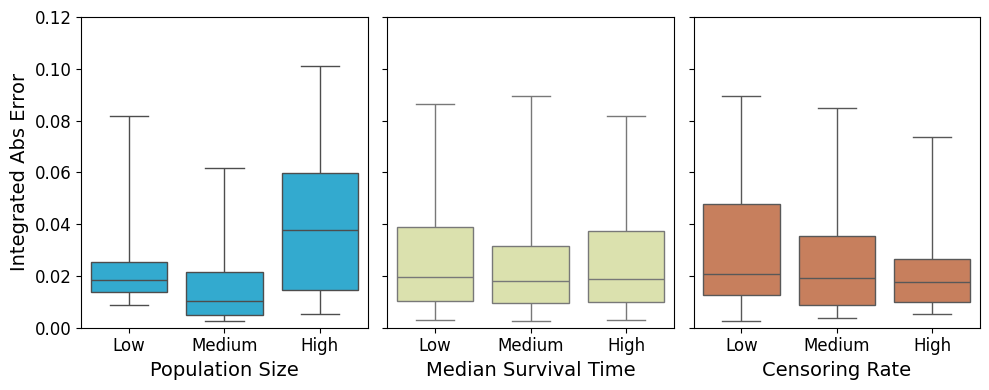}
        \caption{}
        \label{fig:IAEbySubgroup}
    \end{subfigure}

    \caption{Boxplots of Integrated Absolute Error (IAE) across various simulation scenarios and subgroups. (A) Median IAE across all simulation settings. (B) Median IAE grouped by sample size, median survival time, and censoring rate (low, median, high) in the simulation.}
    \label{fig:SimulationIAE}
\end{figure}

To evaluate KM-GPT’s performance under varied parameter settings, the IAE was computed for all 27 simulation subgroups defined by combinations of sample size, median survival time, and censoring rate. The results are visualized in Figure~\ref{fig:SimulationIAE}.

Figure~\ref{fig:SimulationIAE}(A) illustrates the IAE distributions for specific parameter combinations, with the subgroups labeled according to variations (low: L, medium: M, high: H) in sample size, median survival time, and censoring rate. Across most subgroups, the median IAE remained consistently low at approximately 0.02, with relatively narrow interquartile ranges (IQRs) and stable performance across varying simulation conditions. However, ten subgroups showed higher IAE values, including LLH, MLL, MHL, HLL, HLM, HLH, HML, HMM, HHL, and HHM. As identified earlier in the AE analyses, the observed performance deterioration in these subgroups is driven by two main factors: sparsity in event observations nearing the end of the survival curves and visual confusion between curve pixels and axis elements. These factors, which increase the variability in survival data reconstruction toward the tail of the curves, result in inflated IAE distributions in these subgroups.
For further illustration of these issues, we provided selected examples of subgroups with elevated IAE in Supplementary Figure~\ref{fig:bad_recons}. These examples demonstrate how sparsity and pixel-level confusion compound to drive the variability observed in these cases.


Figure~\ref{fig:SimulationIAE}(B) aggregates the IAE results across the three key parameters. As the sample size increased from low to medium, the overall IAE decreased, as larger datasets provided more information to stabilize survival estimates. However, when the sample size became high, performance diminished at the tail of the survival curves, where sparsity in observations amplified "tail effect" inaccuracies. Medium-sized sample sizes offered the most balanced conditions, providing sufficient data resolution to reduce variability without being overly impacted by sparsity in later follow-up periods.

No substantial differences in accuracy were observed across subgroups with varying median survival times.
For censoring rate subgroups, the median IAE values were similar, but high-censoring subgroups exhibited lower variability in IAE compared to low-censoring subgroups. This discrepancy arises because survival curves with minimal censoring frequently extend close to the time axis, where overlapping graphical elements hinder digitization accuracy. This finding underscores an area for future improvements in KM-GPT’s handling of pixel-level features, which could further enhance its robustness in low-censoring scenarios.

\section{Evaluation on Published Clinical KM Plots}
\label{sec:realdata}
In this section, we tested KM-GPT on KM plots obtained from published clinical studies. Unlike the synthetic single-arm KM plots analyzed in the previous section containing only one curve per plot, the KM plots in this evaluation consist of multiple survival curves from comparative clinical studies. These multi-group KM plots introduce greater complexity, reflecting the real-world challenges of curve overlap, group-specific annotations, and diverse figure formats commonly encountered in publications.

We selected KM plots from three metastatic breast cancer (MBC) clinical studies that reported progression-free survival (PFS) or overall survival (OS) outcomes (original plots are displayed in Supplementary Figure~\ref{fig:original_MBC_figures}). These figures were chosen because they reflect the heterogeneity typically found in published KM plots: variations in sample size, follow-up duration, censoring rate, and proportional hazards validity. This diversity provides a robust evaluation setting for KM-GPT. Additionally, these figures had been manually digitized in previous studies~\cite{cancertrialsinfo, anderer2021adaptive} and used with IPDfromKM to reconstruct IPD, serving as a benchmark for comparison.



We first evaluated the performance of KM-GPT 
in extracting survival curve parameters. KM-GPT demonstrated 100\% accuracy in extracting key hyperparameters, including axis information, the number of curves, and the number of individuals at risk from the risk table, as verified through manual validation. In addition, it accurately extracted and matched the group-specific annotation (e.g. treatment vs. comparator arms) extracted from the MMPU module to the reconstructed IPD. 

Next, we assessed the fidelity of KM-GPT’s reconstructed IPD. In the absence of actual individual patient data for direct comparison, we adopted two approaches to evaluate reconstruction accuracy: (1) overlaying the reconstructed survival curves on the original KM plots for direct visual comparison, and (2) comparing the estimated median overall survival (mOS) or median progression-free survival (mPFS) derived from the reconstructed IPD with the reported values from the original studies.

\begin{figure}[htbp!]
    \centering
    
    \begin{subfigure}{0.48\linewidth}
        \centering
        \includegraphics[width=\linewidth]{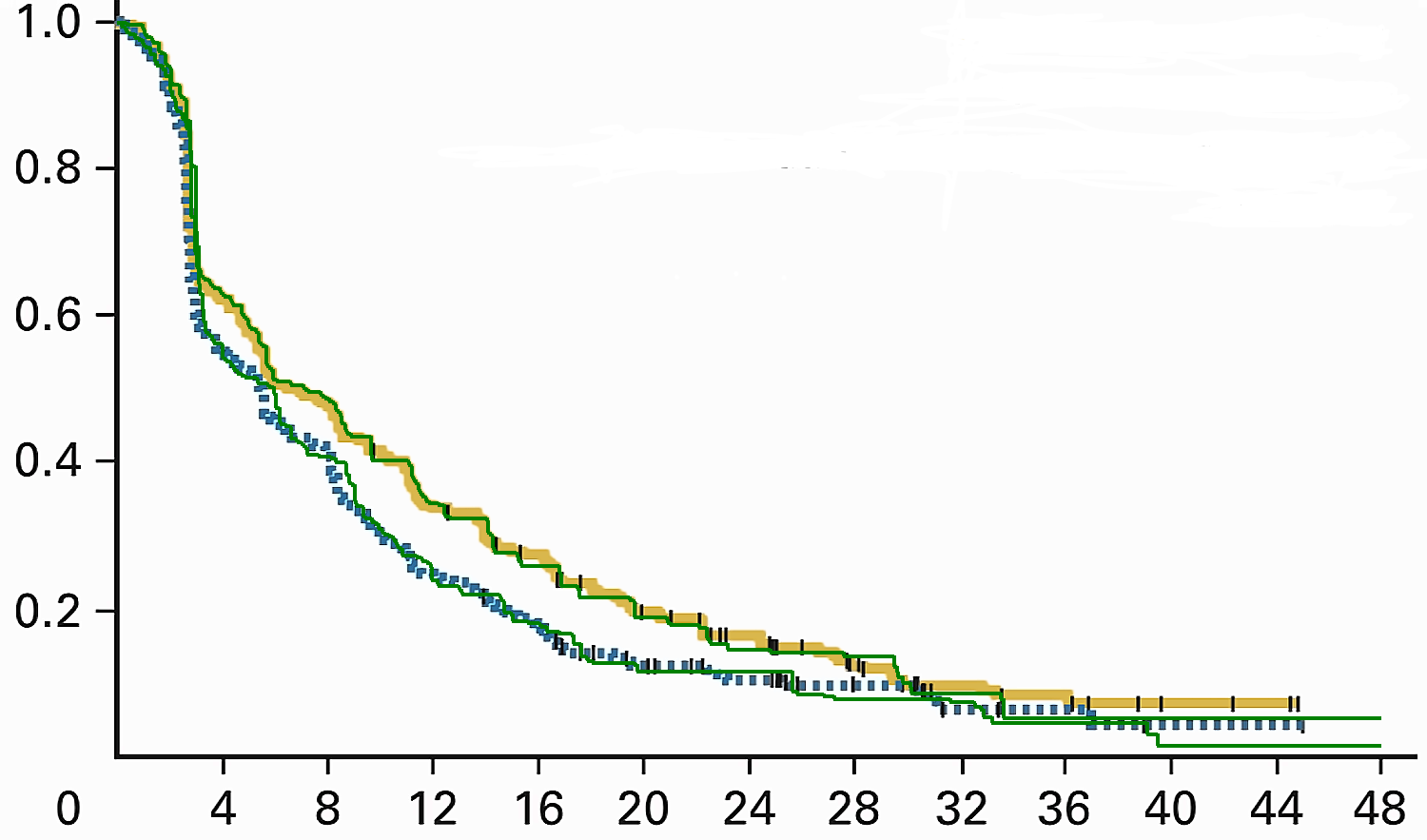}
        \caption*{A, KM-GPT}
        \label{fig:Overlay_KMGPT_1}
    \end{subfigure}
    \hfill
    \begin{subfigure}{0.48\linewidth}
        \centering
        \includegraphics[width=\linewidth]{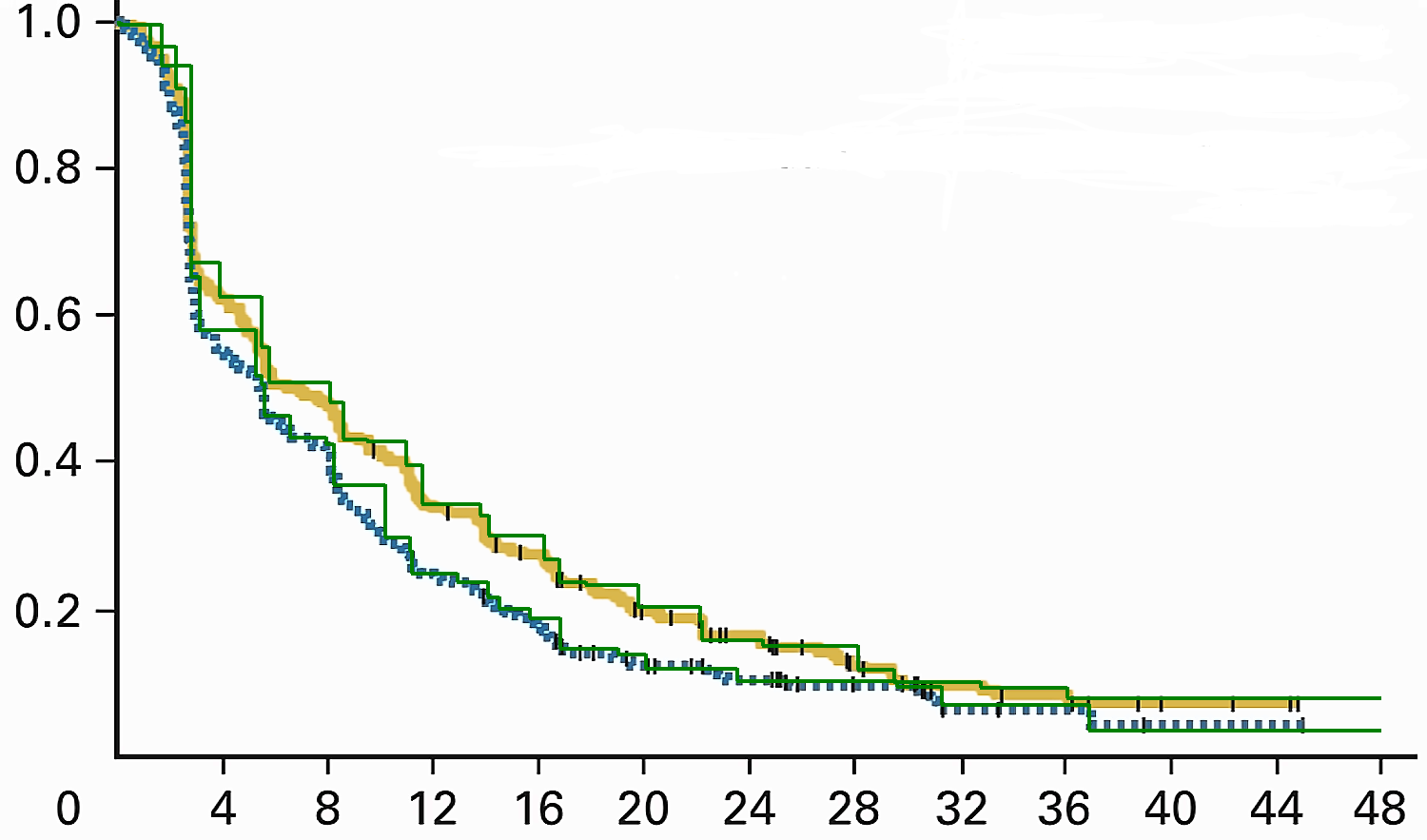}
        \caption*{A, Manual Digitization}
        \label{fig:Overlay_Manual_1}
    \end{subfigure}

    \begin{subfigure}{0.48\linewidth}
        \centering
        \includegraphics[width=\linewidth]{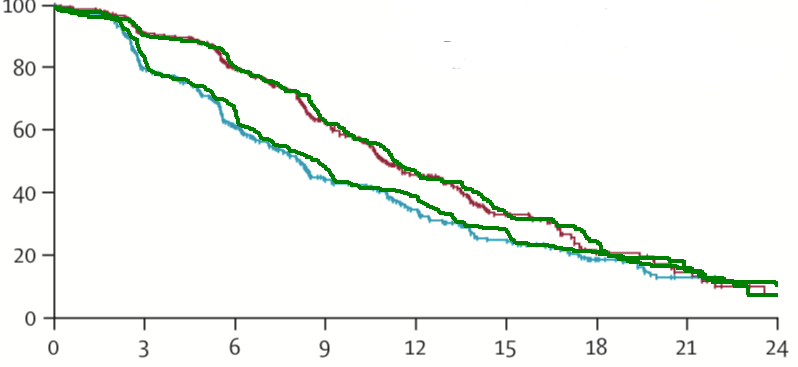}
        \caption*{B, KM-GPT}
        \label{fig:Overlay_KMGPT_2}
    \end{subfigure}
    \hfill
    \begin{subfigure}{0.48\linewidth}
        \centering
        \includegraphics[width=\linewidth]{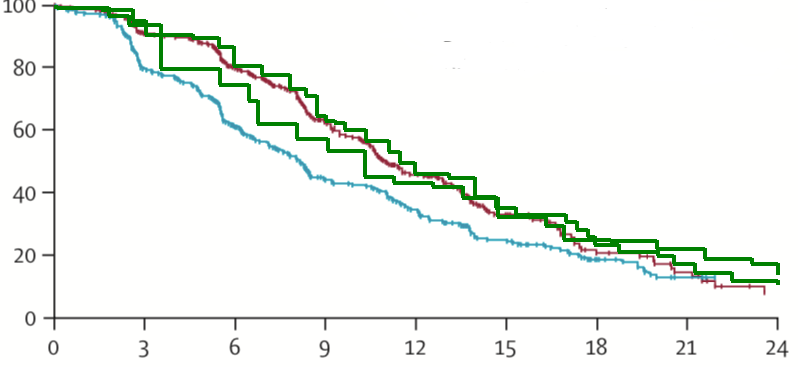}
        \caption*{B, Manual Digitization}
        \label{fig:Overlay_Manual_2}
    \end{subfigure}

    \begin{subfigure}{0.48\linewidth}
        \centering
        \includegraphics[width=\linewidth]{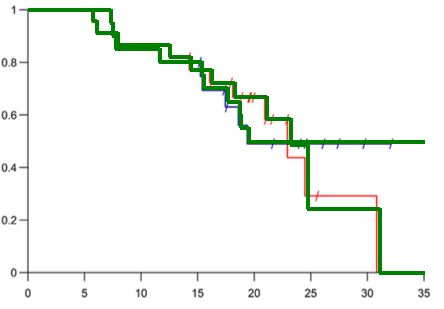}
        \caption*{C, KM-GPT}
        \label{fig:Overlay_KMGPT_3}
    \end{subfigure}
    \hfill
    \begin{subfigure}{0.48\linewidth}
        \centering
        \includegraphics[width=\linewidth]{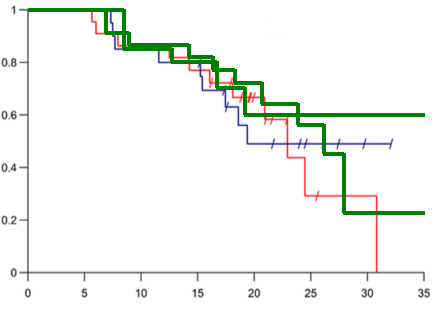}
        \caption*{C, Manual Digitization}
        \label{fig:Overlay_Manual_3}
    \end{subfigure}

    \caption{Reconstructed Survival Curves by KM-GPT and Manual Digitization. Each row corresponds to one trial (A: PMID 23312888, B: PMID 20855825, C: PMID 25892646), with the left column representing KM-GPT and the right column representing manual digitization combined with IPDfromKM. In each panel, reconstructed curves (green lines) are overlaid on the original KM curves.}
    \label{fig:fit-results-overlay-3x2}
\end{figure}

Figure~\ref{fig:fit-results-overlay-3x2} presents a visual comparison of survival curves reconstructed using KM-GPT and manual digitization overlaid on the original KM plots. Across all trials, the curves generated by KM-GPT (green lines, left column) closely aligned with the shape and stepwise dynamics of the published KM estimates, capturing subtle inflections and censoring-driven plateaus. In contrast, manually digitized reconstructions (green lines, right column) exhibited broader deviations, particularly during later follow-up periods when the number at risk was low. This discrepancy was most prominent in the Fulvestrant trials (Panel C), where manual reconstructions overestimated survival probabilities.

Table~\ref{tab:MBC-endpoint} compares the estimated mOS and mPFS values from the reconstructed IPD with the reported values from the original studies, as well as the estimates derived from manual digitization followed by IPDfromKM. KM-GPT’s estimates were close the reported results, including the 95\% CIs in the original studies. The only notable deviation occurred in the upper 95\% CI for the Fulvestrant + Selumetinib arm, where KM-GPT estimated 31.0 months compared to the original report’s value of Not Reached (NR). As shown in Figure~\ref{fig:fit-results-overlay-3x2}, this discrepancy likely stems from the impact of heavy censoring around the median survival time, which affects the accuracy of survival reconstruction and estimation.

In contrast, reconstructions generated from manual digitization showed greater variability and occasionally deviated significantly from the reported values. The largest deviation was observed in the Fulvestrant + Placebo arm, where the reported mOS was 19.4 months, but the manual digitization reconstruction yielded a value of NR.  These findings demonstrate KM-GPT’s ability to deliver high-fidelity survival curve reconstructions that preserve important clinical signals and minimize noise and inconsistencies commonly introduced by manual digitization.

\begin{table}[H]
\centering
\caption{Comparison of Reported Endpoints and Those Estimated from Reconstructed Data. Values are reported as median (95\% CI). mPFS: median progression-free survival; mOS: median overall survival; NR: not reached.}
\label{tab:MBC-endpoint}
\begin{tabular}{lccc}
\toprule
Arm & Reported & KM-GPT & Manually-Digitized \\
\midrule
\multicolumn{1}{l}{\textbf{PMID 23312888}} & \multicolumn{3}{c}{\textbf{Endpoint: mPFS}} \\
\midrule
Bevacizumab + paclitaxel & 11.0 (10.4 - 12.9) & 10.9 (9.8 - 12.2) & 11.1 (10.8 - 13.6) \\
Bevacizumab + capecitabine & 8.1 (7.1 - 9.2) & 8.0 (6.9 - 8.5) & 8.5 (7.5 - 9.4) \\
\midrule
\multicolumn{1}{l}{\textbf{PMID 20855825}} & \multicolumn{3}{c}{\textbf{Endpoint: mPFS}} \\
\midrule
Fulvestrant 500 mg & 6.5 (Not Reported) & 6.8 (5.7 - 7.6) & 8.1 (5.9 - 8.6) \\
Fulvestrant 250 mg & 5.5 (Not Reported) & 5.6 (3.8 - 6.2) & 5.6 (5.2 - 6.6) \\
\midrule
\multicolumn{1}{l}{\textbf{PMID 25892646}} & \multicolumn{3}{c}{\textbf{Endpoint: mOS}} \\
\midrule
Fulvestrant + Selumetinib & 22.9 (16.1 - NR) & 22.9 (16.2 - 31.0) & 23.2 (16.9 - NR) \\
Fulvestrant + Placebo & 19.4 (15.2 - NR) & 19.5 (15.4 - NR) & NR (16.2 - NR) \\

\bottomrule
\end{tabular}
\end{table}

These results also highlight an important limitation of relying solely on hazard ratios (HRs) as summary statistics in clinical trial reporting. HRs assume proportional hazards, an assumption often violated in real-world settings, as demonstrated in Panels B and C of Figure~\ref{fig:fit-results-overlay-3x2}, where the treatment and control curves visibly cross or diverge at different time intervals. In such scenarios, HRs provide only an average effect over the entire follow-up period, potentially obscuring time-varying treatment benefits or risks. In contrast, reconstructed IPD preserves the full shape of the survival curve, enabling more detailed analyses such as time-dependent hazard estimation, restricted mean survival time (RMST) comparisons, and flexible meta-analytic modeling. These capabilities highlight the added value of KM-GPT: not only does it recover high-fidelity survival trajectories, but it also facilitates advanced downstream analyses that extend beyond the constraints of the proportional hazards framework.

\section{Utility of KM-GPT for Downstream Analyses: Meta-Analysis}
\label{sec:meta}
The reconstructed IPD from KM plots generated by KM-GPT opens up numerous opportunities for downstream analyses, significantly expanding the scope of information derivable from published survival data. These applications include meta-analyses across multiple clinical trials to robustly pool survival outcomes, designing new trials by leveraging historical controls to supplement concurrent trials for rare diseases, conducting external validation of predictive models, and performing detailed subgroup analyses at a more granular level. In this section, we present a specific use case: utilizing reconstructed IPD for meta-analysis that does not rely on proportional hazard assumptions. By reconstructing IPD with KM-GPT, we can harmonize and analyze survival data across studies with greater depth, providing richer insights into treatment effects under different subgroup populations.

Immune checkpoint inhibitors (ICIs) targeting the programmed cell death protein 1 (PD-1) receptor and its ligand, programmed death ligand 1 (PD-L1),  play a crucial role in immunotherapy for solid tumors. PD-L1, a transmembrane protein expressed on tumor and immune cells, binds to PD-1 on activated T lymphocytes, suppressing antitumor immune responses and enabling tumor immune evasion. In gastric and gastroesophageal junction (GEJ) cancers, PD-L1 expression, typically measured using the combined positive score (CPS), is a key biomarker for predicting response to ICIs. To better understand how the degree of PD-L1 expression influences ICI efficacy, we focused on three landmark clinical trials: KEYNOTE-061 \citep{shitara2018pembrolizumab}, KEYNOTE-062 \citep{shitara2020efficacy}, and JAVELIN Gastric 100 \cite{moehler2021phase}, which evaluated ICIs in patients with advanced gastric/GEJ cancer. KEYNOTE-061 compared pembrolizumab with paclitaxel in the second-line setting, KEYNOTE-062 evaluated pembrolizumab $\pm$ chemotherapy against chemotherapy alone as first-line treatment, and JAVELIN Gastric 100 studied avelumab as maintenance therapy following first-line chemotherapy. In this meta-analysis, we aimed to assess the pooled efficacy (e.g., mOS) of ICIs across these studies, with particular focus on PD-L1 biomarker-defined subgroups (CPS $\geq$ 1 and CPS $\geq$ 10).


We first evaluated the accuracy of IPD reconstruction through visual inspection of overlaid curves. Supplementary Figure~\ref{fig:PD-L1-Overlay} displays the six original KM plots, with data stratified by PD-L1 expression subgroups (CPS $\geq$ 1 and CPS $\geq$ 10). Each trial contributed two KM plots, corresponding to these biomarker-defined subpopulations. Using KM-GPT, we reconstructed IPD data from the treatment arms of pembrolizumab in KEYNOTE-061 and KEYNOTE-062, and avelumab in JAVELIN Gastric 100, plotted as green lines overlaid on the original curves. The reconstructed curves showed close alignment with the original KM curves across all six treatment arms for ICIs.

To quantitatively evaluate KM-GPT’s accuracy, we computed the mOS with 95\% CIs for the six subgroups and compared these values to the reported mOS results from the original publications (Supplementary Table~\ref{tab:MetaOS}).  Across all trials and biomarker-defined subgroups, the mOS values reconstructed by KM-GPT closely aligned with the reported results. A slightly larger discrepancy was observed in the JAVELIN Gastric 100 study for the CPS $\geq$10 subgroup, where a substantial proportion of patients were censored near the median survival time (Panel F of Supplementary Figure~\ref{fig:PD-L1-Overlay}), reducing the number of observed events available to anchor the reconstruction. This led to a flat segment of the survival curve (approximately from 6 to 10 months), increasing both uncertainty in reconstruction and the vulnerability of endpoint estimation. 


Based on the reconstructed IPD, we conducted a meta-analysis of survival data from the three trials, stratified by PD-L1 expression subgroups. To model survival outcomes, we employed a piecewise exponential framework within a Bayesian hierarchical model. The time axis was partitioned into $J$ disjoint intervals $I_j = (t_{j-1}, t_j]$ for $j=1,\dots,J$, within which the hazard function was assumed to remain constant. The hazard rate for study $s$ during interval $j$ was modeled as constant, with the log-hazard parameter $\alpha_{sj}$ and hazard rate $\lambda_{sj} = \exp(\alpha_{sj})$. Each study-specific log-hazard $\alpha_{sj}$ is modeled as a Gaussian deviation from a pooled log-hazard $a_j$: $
    \alpha_{sj}\sim \mathcal{N}(a_j, \sigma_j^2)$,
where $a_j$ is the pooled effect representing the overall log-hazard in interval $j$, and $
\sigma_j^2$ captures between-study variability. Within this framework, each study is allowed to have its own piecewise constant hazard trajectory, governed by the interval-specific parameters $\alpha_{sj}$. The pooled effect across studies, represented by $a_j$, provides meta-analytic inference that accounts for variation at the study level while estimating a common underlying hazard structure. Full details of the model formulation are provided in Supplementary Materials Section~\ref{sec:meta_model}.

\begin{figure}[htbp!]
    \centering
    \includegraphics[width=0.95\linewidth]{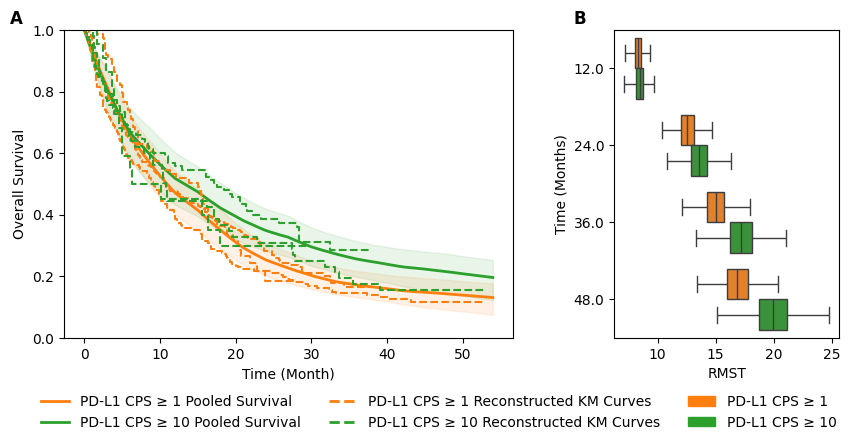}
    \caption{Meta Analysis Stratified by PD-L1 Status. 
    (A) Posterior survival curves for pembrolizumab and avelumab trials, 
    stratified by CPS $\geq 1$ and CPS $\geq 10$ differentiated by colors, with solid lines representing the estimated OS curves and shaded regions indicate 95\% credible intervals from the Bayesian meta-analysis. Curves estimated with reconstructed IPD from three trials stratified by groups are plotted in different line styles. (B) Boxplots of estimated restricted mean survival time (RMST) at 12, 24, 36, and 48 months.}
    \label{fig:PDL1_meta}
\end{figure}

Panel A of Figure~\ref{fig:PDL1_meta} shows the pooled survival curves (solid lines) with 95\% credible intervals (shaded regions). The meta-analytic curves balance information across trials, smoothing out trial-specific fluctuations to provide a stable estimate of survival outcomes over time. In later follow-up periods (beyond 30 months), where individual trial curves exhibit greater variability due to censoring, the posterior summary retains precision, yielding credible intervals that appropriately reflect increasing uncertainty. 
Stratification by PD-L1 expression reveals clear differences in immunotherapy efficacy. Patients with CPS $\geq$10 consistently demonstrate improved survival compared to those with  CPS $\geq$1. For endpoint analysis, the CPS $\geq$10 group achieved a higher mOS compared to the CPS $\geq$1 group, with posterior estimated values of 16.2 months (95\% CI: 10.9–21.9) vs. 13.9 months (95\% CI: 11.0–16.9), respectively. 
This stratification effect is further reflected in the estimated restricted mean survival time (RMST), as shown in Panel B of Figure~\ref{fig:PDL1_meta}. Between 12 and 48 months, the RMST distributions for CPS $\geq$10  consistently shifted higher compared to CPS $\geq$1, with more pronounced differences observed at later time points. These analyses demonstrate the critical role of PD-L1 expression as an effect modifier for immunotherapy efficacy in advanced gastric cancer.

In summary, our meta-analysis of reconstructed IPD demonstrates that KM-GPT not only enables the recovery of high-fidelity survival trajectories but also facilitates detailed subgroup analyses. This highlights the potential of automated IPD reconstruction to enhance evidence synthesis, support biomarker-driven treatment strategies, and provide a scalable platform for advancing precision medicine in oncology.

\section{Conclusion and Discussion}
\label{sec:Discussion}
In this work, we developed KM-GPT, the first fully automated end-to-end pipeline for reconstructing IPD from KM survival plots. By integrating advanced image enhancement, multi-modal reasoning, and iterative reconstruction algorithms, KM-GPT achieves high accuracy, robustness, and reproducibility. A key innovation of KM-GPT lies in the MMPU module, which seamlessly combines OCR techniques with GPT-5’s multi-modal reasoning capabilities. This hybrid design enables the automated interpretation of axes, risk tables, and survival curves, allowing the system to process complex and heterogeneous KM plots without requiring manual input or expert calibration. Another significant strength of KM-GPT is its image preprocessing module, which systematically optimizes raw images for downstream analysis through task-specific transformations such as axis calibration, adaptive thresholding for risk tables, and denoising for survival curve reconstruction.  

KM-GPT introduces a novel paradigm for survival data extraction that impacts clinical research and evidence synthesis. By automating the workflow from figure digitization to IPD reconstruction, KM-GPT enables large-scale and robust secondary analyses, including meta-analyses, systematic reviews, and evidence-based decision-making. This capability is particularly valuable in oncology, where biomarker-driven survival analyses are critical for accelerating therapeutic development and advancing precision medicine. Furthermore, KM-GPT empowers researchers across disciplines, including those with limited programming expertise. With its web-based interface and integrated AI assistant, KM-GPT reduces technical barriers, ensuring accessibility for both clinicians and researchers from diverse backgrounds.


There are many exciting directions for the future development of KM-GPT. First, our analyses of synthetic data and real-world clinical trials revealed that reconstruction accuracy is particularly vulnerable in the tail regions of KM curves, especially when the number of observed events declines sharply. Additionally, as observed in the MBC studies and PD-L1 meta-analysis, the estimation of median survival time becomes unstable when the curve flattens near the 50\% survival threshold. In such cases, the stepwise nature of KM curves amplifies small reconstruction deviations, leading to greater variance in median estimates. Addressing this limitation is a priority for future versions of KM-GPT, and we aim to develop improved axis calibration strategies to mitigate the effects of sparse event data and flat segments. Second, we plan to develop more advanced algorithms capable of reliably processing monochromatic KM plots, where group distinctions are indicated by line styles rather than colors. This functionality is critical for making KM-GPT compatible with a broader range of figure formats and publication standards commonly found in clinical trial reports. Finally, an important direction is to enhance automation by enabling direct extraction and processing of KM plots from PDF trial reports. This includes automatic alignment of extracted KM curves with reported study characteristics, such as subgroup labels and relevant eligibility criteria.

\section*{Data Availability Statement}
The data for this study is accessible through the KM-GPT interface, which is freely available for use at \url{https://km-gpt.wse.jhu.edu/}.

\section*{Acknowledgments}
Yanxun Xu is supported in part by National Institute of Health grants R01MH128085 and R01AI197147. 

\bibliographystyle{unsrt}  
\bibliography{references}  

\newpage
\appendix
\input{supplement}

\end{document}

%% file: supplement.tex
\setcounter{page}{1}
\renewcommand{\thesection}{\Alph{section}}
\renewcommand{\thesubsection}{\Alph{section}.\arabic{subsection}}

\renewcommand{\thefigure}{S\arabic{figure}}
\setcounter{figure}{0} 

\renewcommand{\thetable}{S\arabic{table}}
\setcounter{table}{0} 

\clearpage

\begin{center}
    {\LARGE Supplementary Materials for ``KM-GPT: An Automated Pipeline for Reconstructing Individual Patient Data from Kaplan–Meier Plots"}
\end{center}

\section{Supplementary Figures and Tables}

\begin{figure}[ht!]
    \centering
    \renewcommand{\thesubfigure}{\Alph{subfigure}}
    \begin{subfigure}[b]{0.8\linewidth}
        \centering
        \includegraphics[width=\linewidth]{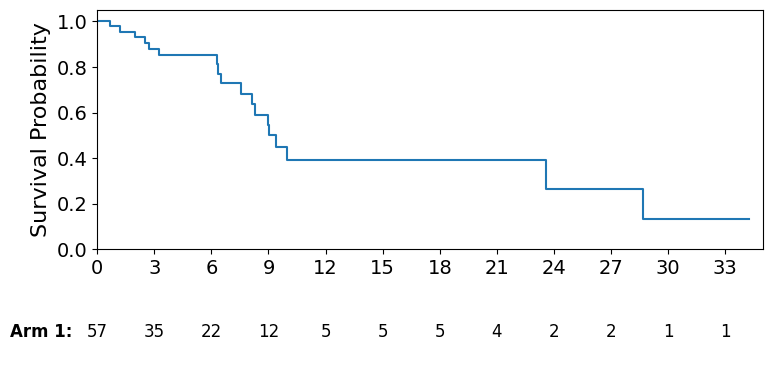}
        \caption{LLH}
        \label{fig:low}
    \end{subfigure}
    \begin{subfigure}[b]{0.8\linewidth}
        \centering
        \includegraphics[width=\linewidth]{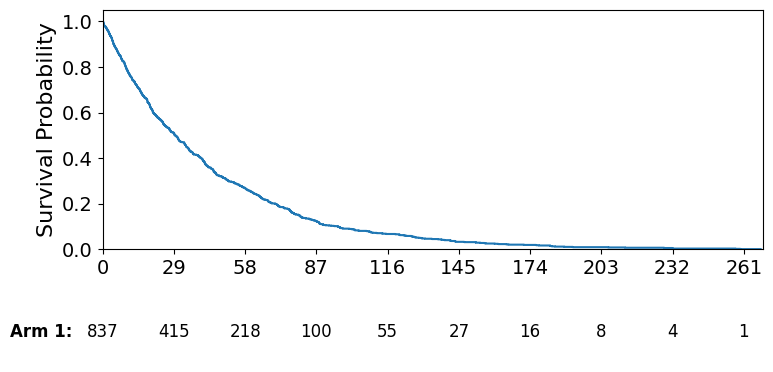}
        \caption{HHL}
        \label{fig:high}
    \end{subfigure}
    \caption{Examples of Simulated Kaplan–Meier Survival Curves. Simulation groups (e.g., LLH and HHL) denote combinations of study sample size, median survival time, and censoring rate, with each letter representing a low (L), medium (M), or high (H) setting for the corresponding parameter.}
    \label{fig:simulated_KM_figures}
\end{figure}

\begin{figure}[ht!]
    \centering
    \includegraphics[width=.95\linewidth]{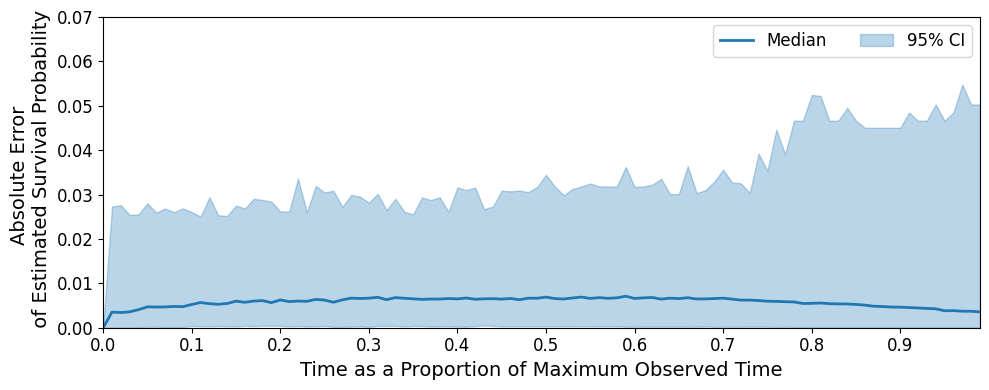}
    \caption{Absolute Error between Reconstructed Survival Curves and Ground Truth across Normalized Time.}
    \label{fig:SimulationAE}
\end{figure}

\begin{table}[ht!]
\centering
\caption{Comparison of Reported and Reconstructed mOS in PD-L1 CPS Subgroups.}
\label{tab:MetaOS}
\begin{tabular}{llcc}
\toprule
Trial & Treatment & Reported mOS & Reconstructed mOS \\
\midrule
\multicolumn{4}{l}{\textbf{PD-L1 CPS $\geq 1$}} \\
\midrule
KEYNOTE-061 & Pembrolizumab & 9.1 (6.2 -- 10.7) & 9.06 (6.23 -- 10.98) \\
KEYNOTE-062 & Pembrolizumab & 10.6 (7.7 -- 13.8) & 10.72 (8.40 -- 14.28) \\
JAVELIN Gastric 100 & Avelumab & 14.9 (8.7 -- 17.3) & 14.97 (8.32 -- 17.26) \\
\midrule
\multicolumn{4}{l}{\textbf{PD-L1 CPS $\geq 10$}} \\
\midrule
KEYNOTE-061 & Pembrolizumab & 10.4 (5.9 -- 18.3) & 10.75 (6.11 -- 17.77) \\
KEYNOTE-062 & Pembrolizumab & 17.4 (9.1 -- 23.1) & 17.52 (9.08 -- 22.12) \\
JAVELIN Gastric 100 & Avelumab & 8.2 (3.9 -- NR) & 6.31 (3.94 -- 18.00) \\
\bottomrule
\end{tabular}
\end{table}

\clearpage

\begin{figure}[ht!]
    \centering
    \renewcommand{\thesubfigure}{\Alph{subfigure}}
    \begin{subfigure}{0.5\linewidth}
        \centering
        \includegraphics[width=\linewidth]{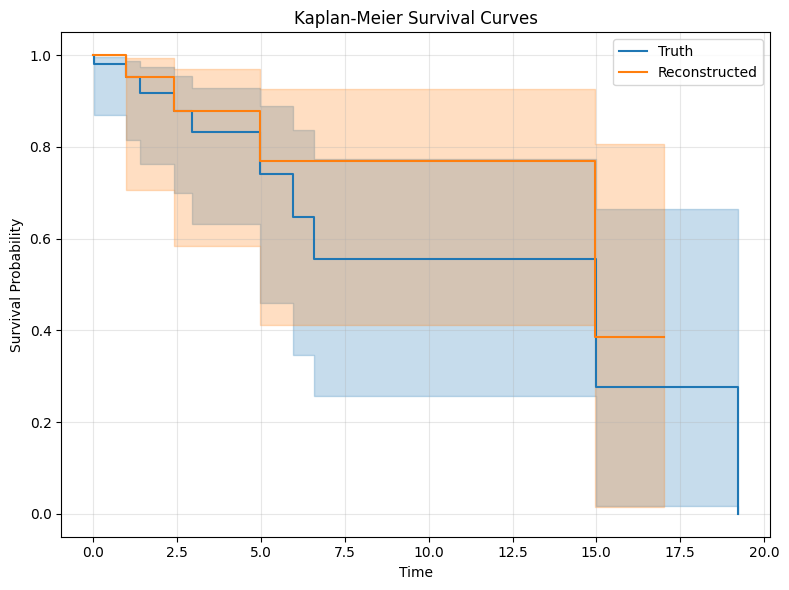}
        \caption{LLH}
        \label{fig:bad_recon_1}
    \end{subfigure}
    \begin{subfigure}{0.5\linewidth}
        \centering
        \includegraphics[width=\linewidth]{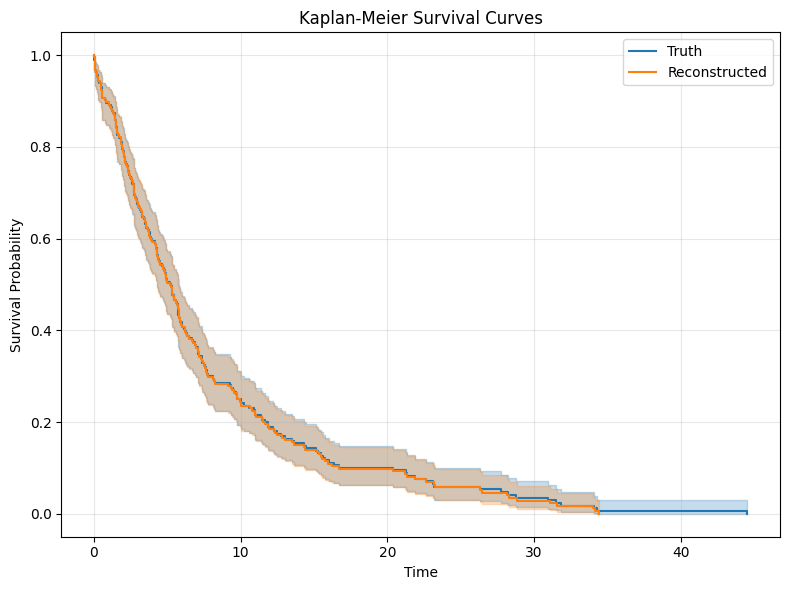}
        \caption{MHL}
        \label{fig:bad_recon_2}
    \end{subfigure}
    \begin{subfigure}{0.5\linewidth}
        \centering
        \includegraphics[width=\linewidth]{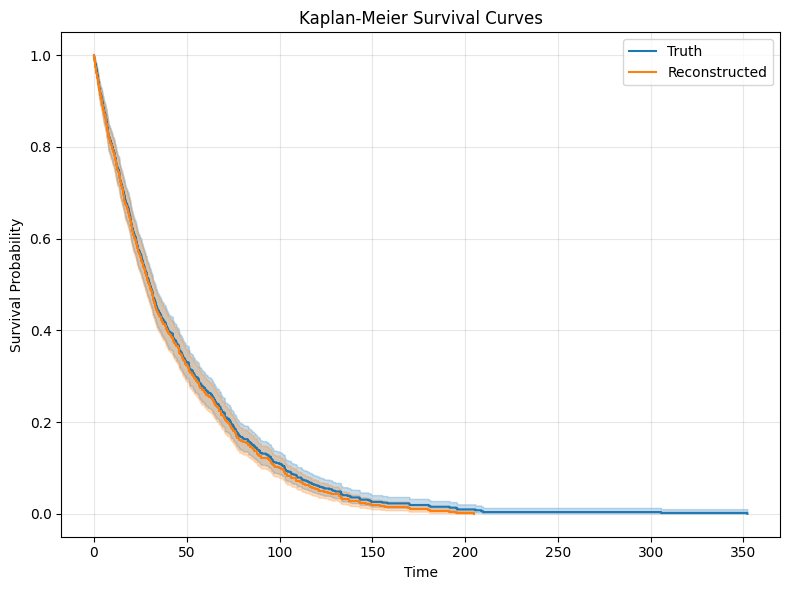}
        \caption{HHL}
        \label{fig:bad_recon_3}
    \end{subfigure}

    \caption{Representative Examples of Poor Reconstructions from Low-Performance Groups. Panel A (LLH) shows a small-sample, high-censoring scenario where steep drops and sparse events in the later follow-up period result in unstable curve estimation. Panels B (MHL) and C (HHL) illustrate a large study with a low censoring rate, where the final portion of the curve is truncated.}
    \label{fig:bad_recons}
\end{figure}

\begin{figure}[ht!]
    \centering
    \renewcommand{\thesubfigure}{\Alph{subfigure}}
    \begin{subfigure}[b]{0.6\linewidth}
        \centering
        \includegraphics[width=\linewidth]{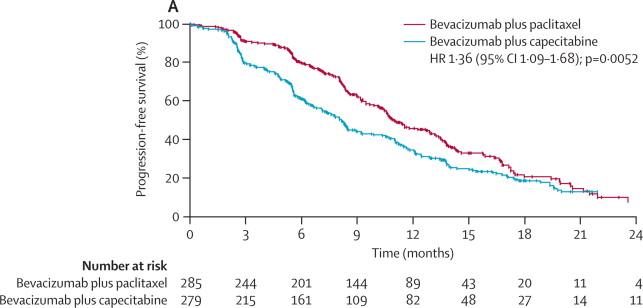}
        \vspace{0.1cm} 
        \caption{PMID 23312888}
        \label{fig:real_KM_1}
    \end{subfigure}
    \begin{subfigure}[b]{0.6\linewidth}
        \centering
        \includegraphics[width=\linewidth]{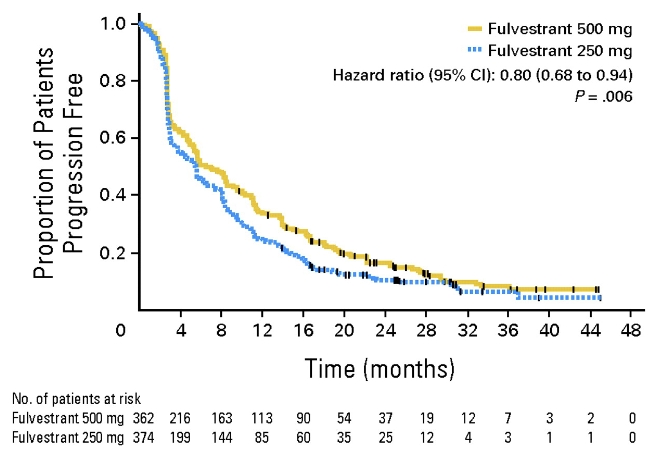}
        \caption{PMID 20855825}
        \label{fig:real_KM_2}
    \end{subfigure}
    \begin{subfigure}[b]{0.6\linewidth}
        \centering
        \includegraphics[width=\linewidth]{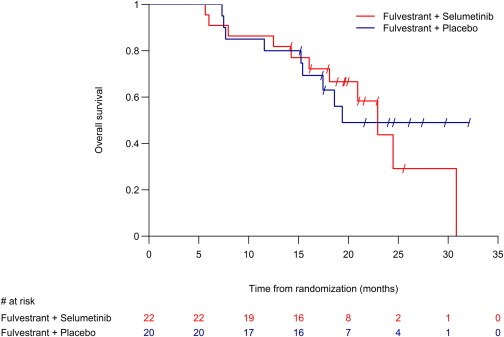}
        \caption{PMID 25892646}
        \label{fig:real_KM_3}
    \end{subfigure}
    \caption{Original KM Plots from Three Metastatic Breast Cancer Studies. }
    \label{fig:original_MBC_figures}
\end{figure}

\begin{figure}[htbp!]
    \centering
    \renewcommand{\thesubfigure}{\Alph{subfigure}}
    \begin{subfigure}{0.45\linewidth}
        \includegraphics[width=\linewidth]{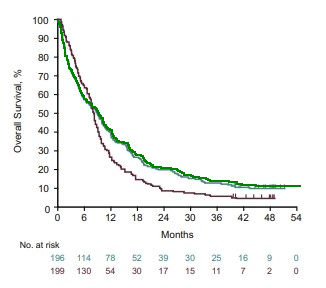}
        \caption{KEYNOTE-061 (CPS $\geq$ 1)}
    \end{subfigure}
    \begin{subfigure}{0.45\linewidth}
        \includegraphics[width=\linewidth]{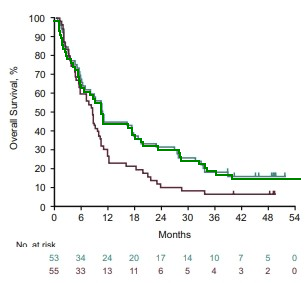}
        \caption{KEYNOTE-061 (CPS $\geq$ 10)}
    \end{subfigure}
    \begin{subfigure}{0.47\linewidth}
        \includegraphics[width=\linewidth]{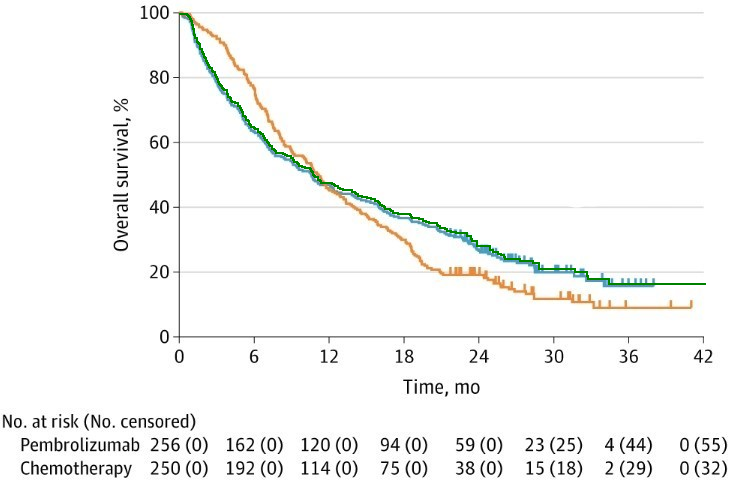}
        \caption{KEYNOTE-062 (CPS $\geq$ 1)}
    \end{subfigure}
    \begin{subfigure}{0.4\linewidth}
        \includegraphics[width=\linewidth]{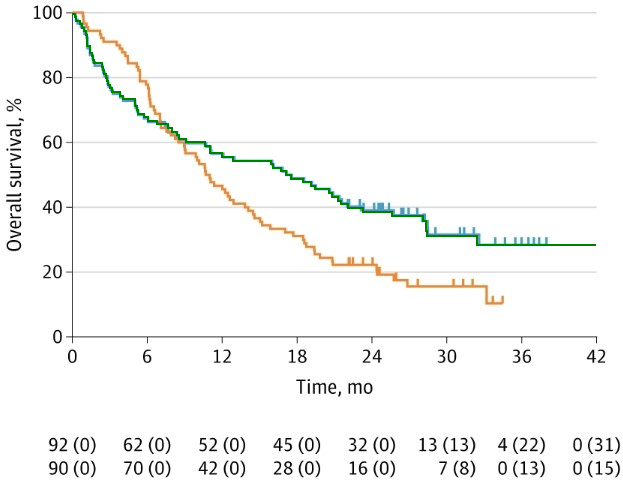}
        \caption{KEYNOTE-062 (CPS $\geq$ 10)}
    \end{subfigure}
    \begin{subfigure}{0.45\linewidth}
        \includegraphics[width=\linewidth]{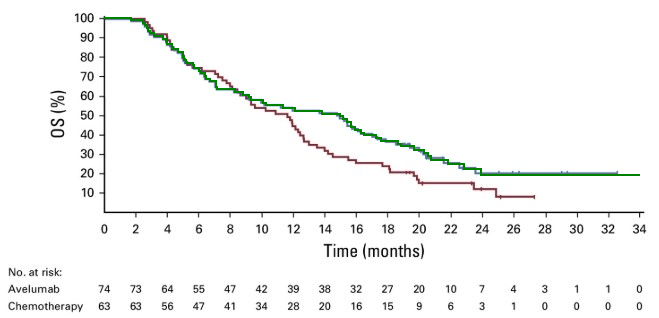}
        \caption{JAVELIN Gastric 100 (CPS $\geq$ 1)}
    \end{subfigure}
    \begin{subfigure}{0.45\linewidth}
        \includegraphics[width=\linewidth]{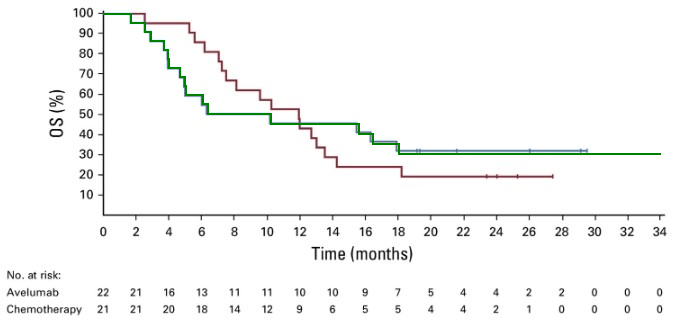}
        \caption{JAVELIN Gastric 100 (CPS $\geq$ 10)}
        \label{fig:mOS_error_fig}
    \end{subfigure}

    \caption{Original KM Plots from Trials with Reconstructed KM Overlaid. Green lines are the reconstructed KM curves from KM-GPT overlaid on ICI treatment arms from original figures.}
    \label{fig:PD-L1-Overlay}
\end{figure}

\begin{figure}[ht!]
    \centering
    \begin{subfigure}[b]{0.45\linewidth}
        \centering
        \includegraphics[width=\linewidth]{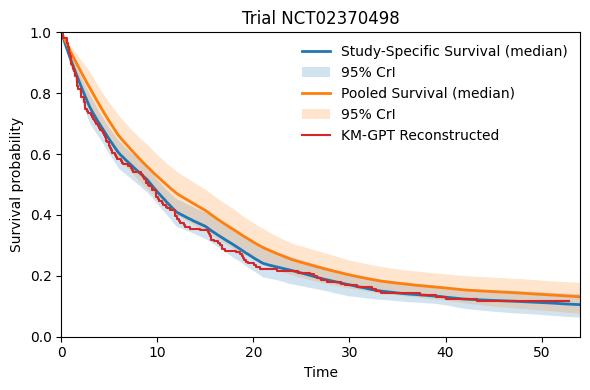}
        \caption{}
        \label{fig:post_1_1}
    \end{subfigure}
    \begin{subfigure}[b]{0.45\linewidth}
        \centering
        \includegraphics[width=\linewidth]{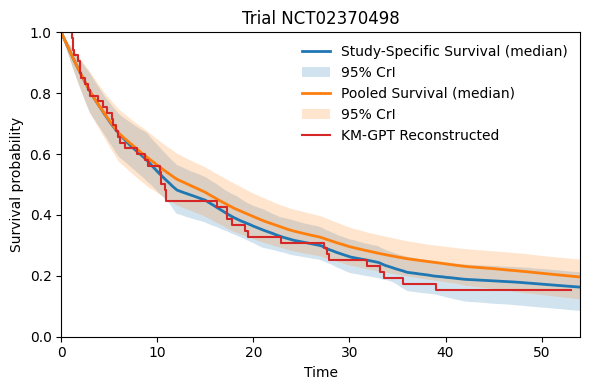}
        \caption{}
        \label{fig:post_10_1}
    \end{subfigure}

    \begin{subfigure}[b]{0.45\linewidth}
        \centering
        \includegraphics[width=\linewidth]{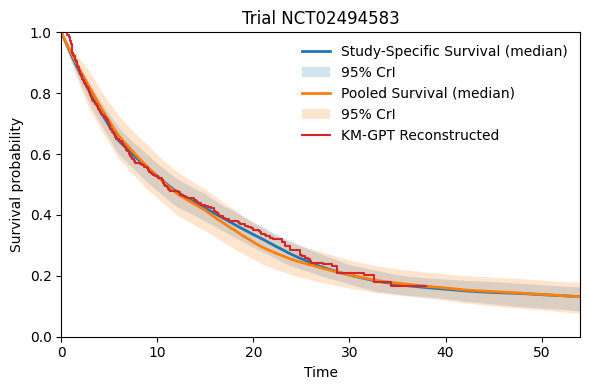}
        \caption{}
        \label{fig:post_1_2}
    \end{subfigure}
    \begin{subfigure}[b]{0.45\linewidth}
        \centering
        \includegraphics[width=\linewidth]{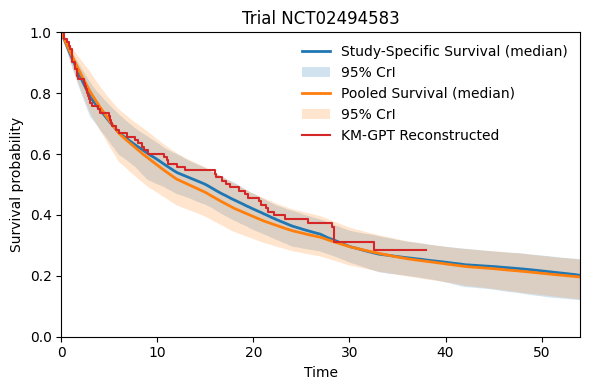}
        \caption{}
        \label{fig:post_10_2}
    \end{subfigure}
    
    \begin{subfigure}[b]{0.45\linewidth}
        \centering
        \includegraphics[width=\linewidth]{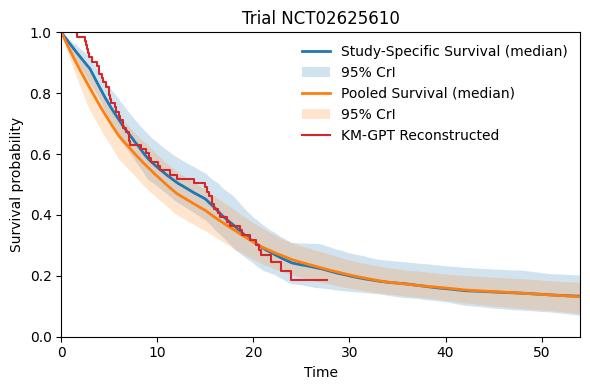}
        \caption{}
        \label{fig:post_1_3}
    \end{subfigure}
    \begin{subfigure}[b]{0.45\linewidth}
        \centering
        \includegraphics[width=\linewidth]{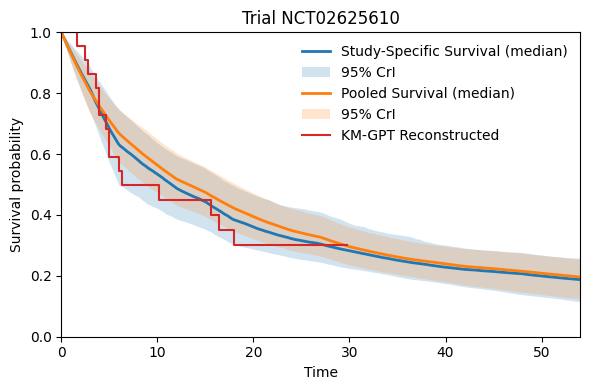}
        \caption{}
        \label{fig:post_10_3}
    \end{subfigure}

    \caption{Posterior Inference of Study-Specific and Pooled Survival Curves.}
    \label{fig:post_grid}
\end{figure}

\clearpage
\section{OCR Engine Settings}
\label{sec:OCR_Engine_setting}
To optimize text extraction from Kaplan–Meier plots, we configured the OCR engine with the following parameters:

\begin{itemize}
  \item \textbf{OCR Engine Mode:} KM-GPT utilizes the \textbf{oem 3} mode of the OCR engine, which leverages an advanced LSTM-based model for text recognition. This mode is highly accurate across a variety of fonts and image qualities, ensuring reliable text extraction.
  \item \textbf{Risk Table Loading Mode}: For loading risk tables, we set the engine to \textbf{psm 6}, a mode specifically designed for extracting structured content, such as rows and columns, from number-at-risk tables.

\end{itemize}

Combining \textbf{oem 3} for high character recognition accuracy and \textbf{psm 6} for structured data extraction ensures the robust and consistent parsing of layout-dependent textual information in Kaplan–Meier plots. These preprocessing steps form the foundation for the Multi-Modality Processing Unit (MMPU), which further refines and interprets OCR outputs using GPT-5 to achieve high-fidelity table reconstruction.

\section{KM-GPT Technical Details}
\label{sec:Method Details}





\subsection{Axis Calibration}
\label{sec:axis_cal}
Following preprocessing, axis calibration is performed to map image pixels into real-valued time and survival probability domains. Tick labels extracted via OCR are parsed by the \textit{range detection} routine, which infers four key parameters: $t_{\min}, t_{\max}, s_{\min}, s_{\max}$. Unique numeric labels are then sorted, and pairwise gaps between labels are computed. The most common increment is identified from a trimmed histogram of these differences, forming the time increment $\Delta_t$ and survival probability increment $\Delta_s$.   For irregular or non-monotonic sequences, the median of the increments is used as the $\Delta$ value. 

Axis endpoints and orientations are determined by the axis identification procedure, which projects ink densities along columns and rows of the lightness channel to locate vertical and horizontal axis strokes. From these projections, the pixel coordinates of the axis baselines $(u_{x_0}, u_{x_1})$ and $(v_{y_0}, v_{y_1})$ are established, and interior margins are automatically adjusted based on surrounding whitespace gradients. 

Finally, an affine transformation is applied to convert pixel coordinates $(u,v)$ into real-world values using the equations:
\[
t(u) = t_{\min} + \frac{u - u_{x_0}}{u_{x_1} - u_{x_0}} \,(t_{\max} - t_{\min}), 
\qquad
s(v) = s_{\max} - \frac{v - v_{y_0}}{v_{y_1} - v_{y_0}} \,(s_{\max} - s_{\min}),
\]
where the inversion of $s(v)$ accounts for the image origin being in the top-left corner. This calibration ensures sub-pixel accuracy in quantifying digitized survival curves. The same transformations are also used to convert curve pixels into their corresponding time and survival probabilities.

\subsection{Curves Differentiation}
\label{sec:curve_diff}
To partition foreground pixels into curve-specific groups, we use a color-space partitioning approach. For each pixel, features $(h,s,l)$ in HSL color space are extracted and stored in a DataFrame. Near-background pixels are optionally removed by retaining only those with lightness  $l \ge 0.2$, effectively suppressing pale grid-lines and page artifacts. The features are then standardized for clustering: if image enhancement is applied, the $s$ (saturation) and $l$ (lightness) channels are up-weighted by a factor of 100 to amplify chroma and luminance differences relative to the $h$ channel. 
We fit a $K$-medoids model~\cite{schubert2019faster} with $K= \text{Num of Curves}$ using Euclidean distance in $(h,s,l)$. The best model, determined by minimizing inertia, is selected, and pixels are assigned to clusters based on the nearest medoids, producing distinct curve labels. The result is a grouped pixel set, where each color represents one unique curve.

\subsection{Overlapping Curve Interpolating}
\label{sec:overlap_curve}
To address overlapping segments on curves, we implement a local $k$-NN consensus scoring method followed by a grid-constrained path tracing technique. Pixels are first sorted and embedded in a Euclidean $k$-NN graph~\cite{scikit-learn}, where the labels of neighboring pixels are analyzed. For each pair of neighbors $i$ and $j$, we construct a same-group indicator matrix $\mathbb{I}_{ij}\in\{-1,+1\}$,  where $+1$ indicates the neighbor $j$ shares $i$'s cluster label and $-1$ otherwise. Neighbor contributions are weighted using inverse-squared distance,  $w_{ij}=1/(d_{ij}^{2}+\varepsilon)$, where  $\varepsilon=10^{-10}$ ensures numerical stability for very small separations. The per-pixel consensus score, estimating local label confidence in overlapping or parallel regions, is computed as the normalized weighted sum: 
$\,\mathrm{score}_i=\frac{1}{k}\sum_{j} \mathbb{I}_{ij}\,w_{ij}.$
Using these consensus scores, we trace the curves and interpolate points in the overlapping regions, ensuring accurate segment reconstruction where curves overlap or run closely together.

\section{Hierarchical Piecewise-Exponential Model for Meta Analysis}
\label{sec:meta_model}
\subsection{Model Formulation}
In this section, we present the Bayesian hierarchical model for meta-analysis. The time axis is partitioned into $J$ disjoint intervals, $I_j = (t_{j-1}, t_j]$ for $j=1,\dots,J$, within which the hazard function for each study is assumed to be constant.

The hazard rate for study $s$ in interval $j$ is $\lambda_{sj} = \exp(\alpha_{sj})$, where $\alpha_{sj}$ is the study- and interval-specific log-hazard. To pool information across studies, each $\alpha_{sj}$ is modeled as a Gaussian deviation from a pooled, interval-specific log-hazard $a_j$:
\begin{equation}
\alpha_{sj} \mid a_j, \sigma_j^2 \sim \mathcal{N}(a_j, \sigma_j^2).
\end{equation}
Here, $a_j$ represents the overall meta-analytic log-hazard in interval $j$, and $\sigma_j^2$ captures the between-study heterogeneity for that interval.

To share strength across adjacent time intervals and ensure the pooled hazard evolves smoothly, we place a hierarchical prior on the sequence of pooled parameters $\mathbf{a} = (a_1, \ldots, a_J)$. Specifically, we model $a_j$ as varying around a latent mean $\mu_j$:
\begin{equation}
a_j \mid \mu_j, \sigma_a^2 \sim \mathcal{N}(\mu_j, \sigma_a^2),
\end{equation}
where the latent process $\boldsymbol{\mu} = (\mu_1, \ldots, \mu_J)$ follows a stationary autoregressive process of order one (AR(1)) to enforce smoothness:
\begin{align*}
\mu_1 \mid \phi, \tau^2 &\sim \mathcal{N}\left(0, \frac{\tau^2}{1-\phi^2}\right), \\
\mu_j \mid \mu_{j-1}, \phi, \tau^2 &\sim \mathcal{N}\left(\phi \mu_{j-1}, \tau^2\right), \quad \text{for } j=2,\ldots,J,
\end{align*}
with $|\phi| < 1$ ensuring stationarity and $\tau > 0$. 

We complete the model specification with the following prior distributions. The standard deviation parameters $\sigma_j$ (characterizing between-study heterogeneity), $\sigma_a$ (governing variation of the pooled effects around the latent mean), and $\tau$ (the innovation standard deviation of the latent process), are assigned weakly informative Half-Normal priors: $\sigma_j \sim \mathcal{N}^+(0, 0.2^2)$, $\sigma_a \sim \mathcal{N}^+(0, 0.2^2)$, and $\tau \sim \mathcal{N}^+(0, 1^2)$. The autoregressive parameter $\phi$ is modeled via a transformed parameter to maintain the stationarity constraint $|\phi| < 1$; specifically, we define $\phi = \tanh(\psi)$ and assign the prior $\psi \sim \mathcal{N}(0, 0.75^2)$.

The pooled hazard in interval $j$ is $\lambda^{\text{pool}}_j = \exp(a_j)$, and for study $s$ the study-specific hazard is $\lambda_{sj} = \exp(\alpha_{sj})$. The corresponding survival functions at time $t$ are derived from the cumulative hazard. Let $\Delta_k = t_k - t_{k-1}$ be the length of the $k$-th interval and let $j(t)$ be the index of the interval such that $t \in I_{j(t)}$.
The survival probability for the pooled population is given by:
\begin{align*}
S_{\text{pool}}(t) &= \exp\left(-\sum_{k=1}^{j(t)} \lambda^{\text{pool}}_k \Delta_k\right),
\end{align*}
and for study $s$, it is given by:
\begin{align*}
S_s(t) &= \exp\left(-\sum_{k=1}^{j(t)} \lambda_{sk} \Delta_k\right).
\end{align*}

Figure~\ref{fig:post_grid} displays the posterior survival trajectories from the hierarchical piecewise-exponential model. Each panel illustrates the reconstructed survival curve, the study-specific curves, and the pooled survival function. Shaded bands represent 95\% credible intervals, derived from 5000 posterior draws after 2000 warm-up steps of Markov Chain Monte Carlo (MCMC) sampling. The study-specific curves capture heterogeneity across trials, reflecting variability in population size, follow-up duration, and censoring patterns. In contrast, the pooled curve provides a stable summary that smooths out trial-level fluctuations. These results demonstrate the model's ability to recover individual trial survival patterns while also producing a coherent pooled estimate that balances study-specific evidence with cross-study information.